\documentclass[11pt]{article}

\usepackage[a4paper,margin=1in]{geometry}
\usepackage{amsmath,amssymb}
\usepackage{graphicx}
\usepackage{booktabs}
\usepackage{array}
\usepackage{float}
\usepackage{subcaption}
\usepackage{hyperref}
\usepackage{xcolor}
\usepackage{longtable}
\usepackage{cite}
\usepackage{caption}

\graphicspath{{../figures/}}

\title{Geometry-Aware R-Structured Kolmogorov--Arnold Networks}

\author{S. Kucherenko, N. Shah \\
Imperial College London, London SW7 2AZ, UK}

\date{\today}

\newcommand{\Rrect}{R_{\mathrm{rect}}}

\begin{document}
\maketitle

\begin{abstract}

We propose a novel hybrid neural architecture, the Geometry-aware R-Structured Kolmogorov-Arnold Network (GRS-KAN), which integrates V.~L.~Rvachev's R-functions into the Kolmogorov-Arnold Network (KAN) framework. The proposed approach combines two complementary modeling mechanisms: smooth nonlinear structure is learned by KAN branches, while known geometric or logical constraints are encoded analytically using differentiable R-functions. This enables explicit representation of discontinuities, feasible regions, and implicit geometric boundaries within a trainable neural architecture.

The framework implements differentiable logical operations through R-conjunctions and R-disjunctions, allowing complex geometric supports to be represented analytically and incorporated directly into regression models. Several GRS-KAN variants are introduced, including additive, multiplicative, and agnostic branch-weighted architectures.

The method is demonstrated on regression problems involving discontinuities with circular and rectangular supports. Numerical experiments show that explicit geometric encoding substantially improves predictive accuracy and boundary localization compared with standard KANs. In the considered benchmarks, geometry-aware GRS-KAN models reduce test RMSE by up to 67\% while simultaneously improving interpretability through explicit analytical representation of the learned geometric structure. The agnostic variant further demonstrates the ability to automatically determine whether geometric priors are beneficial for a given learning task.

\end{abstract}

\section{Introduction}

Neural networks have achieved remarkable success across numerous domains, yet their ``black box'' nature remains a significant limitation for applications requiring interpretability and formal verification, particularly in safety-critical fields such as pharmaceutical manufacturing, process control, and scientific computing. Two recent developments address this challenge from complementary perspectives: Kolmogorov--Arnold Networks (KANs) and R-functions.

KANs, introduced by Liu et al.~\cite{liu2024kan}, replace traditional weight matrices with learnable univariate functions, offering improved interpretability and accuracy for scientific computing tasks. The architecture is grounded in the Kolmogorov--Arnold representation theorem, which states that any continuous multivariate function can be decomposed into sums of compositions of univariate functions.

The introduction of KANs triggered an exceptionally rapid expansion of research activity. Within approximately one year of publication, the original KAN paper accumulated several thousand citations and generated a rapidly growing ecosystem of derivative architectures and applications. Numerous variants have already appeared, including FastKAN, ConvKAN, GraphKAN, TemporalKAN, PDE-KAN, Physics-Informed KANs, Symbolic-KANs, FourierKAN, WaveletKAN, ChebyshevKAN, Transformer--KAN hybrids, operator-learning KANs, and multiple domain-specific implementations in medical imaging, remote sensing, and segmentation tasks. This rapid diversification reflects the broader interest in replacing conventional node-based neural nonlinearities with learnable edge-based functional representations.

Most existing KAN extensions focus primarily on modifications of approximation mechanisms, including alternative basis functions, spline parameterizations, convolutional operators, symbolic primitives, attention mechanisms, or physics-informed loss constructions. However, comparatively little attention has been devoted to incorporating explicit geometry, Boolean structure, analytical feasible regions, discontinuity-aware compositions, or implicit surface representations directly into the KAN framework.

Separately, R-functions, developed by V.L. Rvachev~\cite{rvachev1982} and extended by Shapiro~\cite{shapiro1991,shapiro2007}, provide a rigorous bridge between logical operations and real analysis. R-functions allow complex geometric shapes and Boolean conditions to be represented analytically by single differentiable equations. This makes them particularly attractive for embedding geometric constraints and logical structure into differentiable machine learning architectures.

This paper introduces the Geometry-aware R-Structured Kolmogorov--Arnold Network (GRS-KAN), a hybrid architecture that embeds R-functions directly into the KAN framework. Conceptually, the proposed framework separates two fundamentally different modeling roles:
\begin{itemize}
\setlength{\itemsep}{-2pt}
\setlength{\parskip}{0pt}
\setlength{\parsep}{0pt}
\item KAN branches learn smooth nonlinear functional structure from data,
\item R-functions encode known geometric or logical structure analytically.
\end{itemize}
This leads to a substantially different direction from existing KAN variants. Rather than introducing another spline modification or approximation basis, the proposed framework incorporates explicit analytical geometry directly into the network architecture itself. The resulting model enables interpretable representation of discontinuities, geometric supports, feasible regions, and implicit boundaries while preserving differentiability required for gradient-based optimization.

The key contributions of this work are:
\begin{enumerate}
\setlength{\itemsep}{-2pt}  
\setlength{\parskip}{0pt}
\setlength{\parsep}{0pt}
\item A theoretical framework for integrating analytical R-function representations of geometric and logical constraints into KANs within differentiable neural architectures
\item Explicit analytical constructions of geometric indicators using R-conjunctions and R-compositions, together with closed-form gradient expressions suitable for backpropagation.
\item Three architectural variants: targeted additive, targeted multiplicative, and agnostic branch-weighted GRS-KANs.
\item Comprehensive numerical experiments demonstrating improved accuracy, geometric localization, and interpretability on problems with known discontinuous structure.
\item Empirical evidence that learnable branch weights can automatically determine whether explicit geometric priors are beneficial for the regression task.
\end{enumerate}

Although the numerical examples in this paper employ simple rectangles and circles, the proposed framework is applicable to arbitrary implicit regions represented analytically by R-functions, including non-convex domains, disconnected regions, and Boolean combinations of multiple constraints.

The remainder of this paper is organized as follows. Section~2 reviews the theoretical background of Kolmogorov--Arnold Networks and R-functions, including the analytical construction of geometric primitives used throughout the paper. Section~3 introduces the proposed GRS-KAN architectures, including additive, multiplicative, and agnostic branch-weighted variants. Section~4 presents numerical experiments on smooth and discontinuous benchmark problems, together with pruning diagnostics, geometric localization studies, and comparisons against standard KAN models. Section~5 discusses the implications of explicit geometry-aware learning and the role of learnable branch selection. Finally, Section~6 summarizes the main conclusions and outlines directions for future research, including extensions to multi-region constraints, adaptive geometric gates, and scientific machine learning applications.

\section{Background and Theory}

\subsection{Kolmogorov--Arnold Networks (KANs)}

Kolmogorov--Arnold Networks (KANs), introduced by Liu et al.~\cite{liu2025kan}, are a class of neural networks motivated by the Kolmogorov--Arnold representation theorem. Unlike conventional Multi-Layer Perceptrons (MLPs), where nonlinear activation functions are attached to nodes and the edges contain scalar weights, KANs place learnable nonlinear functions directly on the edges of the network. This architectural change leads to improved interpretability, sparse compositional structure discovery, and strong approximation performance for scientific machine learning problems.

The theoretical foundation of KANs is the Kolmogorov--Arnold representation theorem, which states that any continuous multivariate function
$f:[0,1]^n \rightarrow \mathbb{R}$ can be represented as a finite superposition of continuous univariate functions:

\begin{equation}
f(\mathbf{x})
=
\sum_{q=0}^{2n}
\Phi_q
\left(
\sum_{p=1}^{n}
\psi_{q,p}(x_p)
\right),
\label{eq:ka_theorem}
\end{equation}
where
$\psi_{q,p}:[0,1]\rightarrow\mathbb{R}$
and
$\Phi_q:\mathbb{R}\rightarrow\mathbb{R}$
are continuous univariate functions. The theorem demonstrates that high-dimensional nonlinear mappings can, in principle, be decomposed into sums and compositions of one-dimensional functions.

KANs implement this idea directly by replacing fixed linear weights with learnable univariate edge functions. A KAN layer maps an input vector
$x_l \in \mathbb{R}^{n_l}$
to an output vector
$x_{l+1}\in\mathbb{R}^{n_{l+1}}$
through

\begin{equation}
x_{l+1,j}
=
\sum_{i=1}^{n_l}
\phi_{l,j,i}(x_{l,i}),
\label{eq:kan_layer}
\end{equation}
where
$\phi_{l,j,i}$
is a trainable univariate function associated with the edge connecting neuron $i$ in layer $l$ to neuron $j$ in layer $l+1$.
Thus, the network performs nonlinear functional transformations directly on the connections rather than on the nodes.

Equation~(\ref{eq:kan_layer}), adopted from the original KAN
formulation of Liu \emph{et al.}~\cite{liu2025kan}, constitutes the
fundamental computational building block of a KAN.
Throughout this paper, this smooth KAN mapping
serves as the baseline function approximator. 
The key idea of the
proposed Geometry-aware R-Structured KAN (GRS-KAN) is not to modify the
internal KAN layer itself, but rather to augment the smooth KAN
approximation with explicit analytical geometric components constructed
from R-functions. Consequently, the KAN branch learns smooth nonlinear
behaviour, while the R-function branch represents known geometric
constraints, discontinuities, feasible regions, and implicit boundaries.

Following Liu et al.~\cite{liu2025kan}, each edge activation is represented using a residual formulation consisting of a smooth base activation plus a spline correction:
\begin{equation}
\phi(x)
=
w_b\, b(x)
+
w_s\, \operatorname{spline}(x),
\qquad
b(x)=\operatorname{silu}(x)
=
\frac{x}{1+\exp(-x)},
\label{eq:kan_activation}
\end{equation}
where $w_b$ and $w_s$ are learnable scalar coefficients. The smooth residual branch provides stable global behaviour, while the spline component captures localized nonlinear structure.

The spline term is parameterized using cubic B-spline basis functions:

\begin{equation}
\operatorname{spline}(x)
=
\sum_i c_i B_i(x),
\label{eq:spline}
\end{equation}
where $B_i(x)$ are cubic B-spline basis functions and $c_i$ are trainable spline coefficients optimized by backpropagation.

Compared with MLPs, KANs exhibit several important properties:

\begin{itemize}
\item \textbf{Interpretability:}
Since nonlinearities are attached to edges as explicit one-dimensional functions, the learned transformations can be visualized directly and, in some cases, symbolically simplified.

\item \textbf{Compositional sparsity:}
Many scientific functions possess low-dimensional compositional structure. KANs naturally expose this through pruning and symbolic snapping of weak or redundant branches.

\item \textbf{Improved scaling laws:}
Liu et al.~\cite{liu2025kan} demonstrated empirically that KANs can achieve higher accuracy than comparable MLPs with fewer parameters on scientific regression tasks.

\item \textbf{Symbolic recovery:}
For several benchmark functions, pruned KANs recover exact analytical expressions after symbolic simplification.
\end{itemize}

A key feature of KANs is structural pruning. During training, many edge activations become negligible, allowing hidden nodes and edges to be removed without degrading accuracy. This often reveals compact compositional representations of the target function.

\begin{figure}[H]
  \hspace*{-0.05\linewidth}
  \begin{minipage}{1.1\linewidth}

    \begin{subfigure}{\linewidth}
      \centering
      \includegraphics[width=\linewidth]{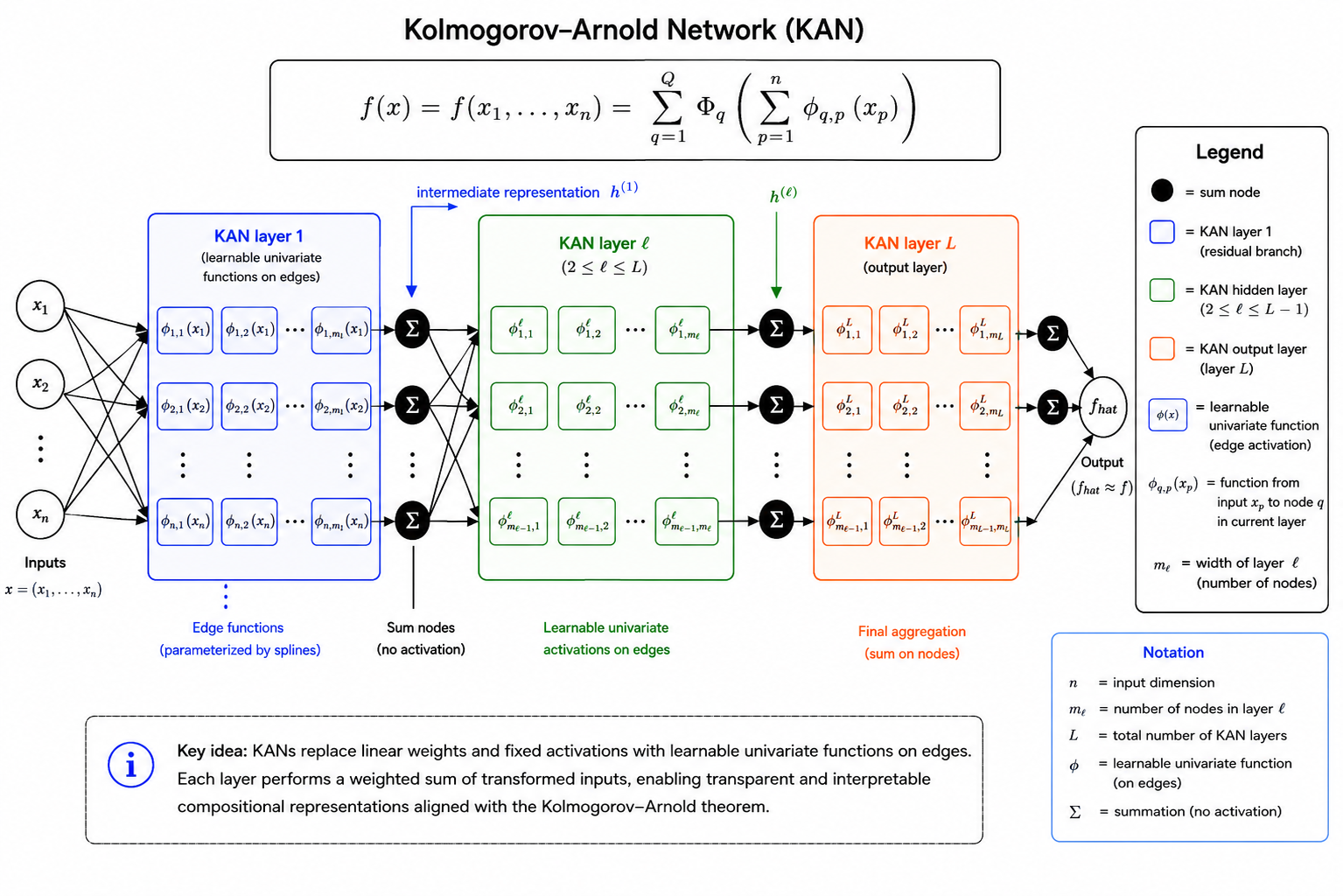}
      \caption{Standard KAN architecture.}
    \end{subfigure}

    \vspace{1ex}

    \begin{subfigure}{\linewidth}
      \centering
      \includegraphics[width=\linewidth]{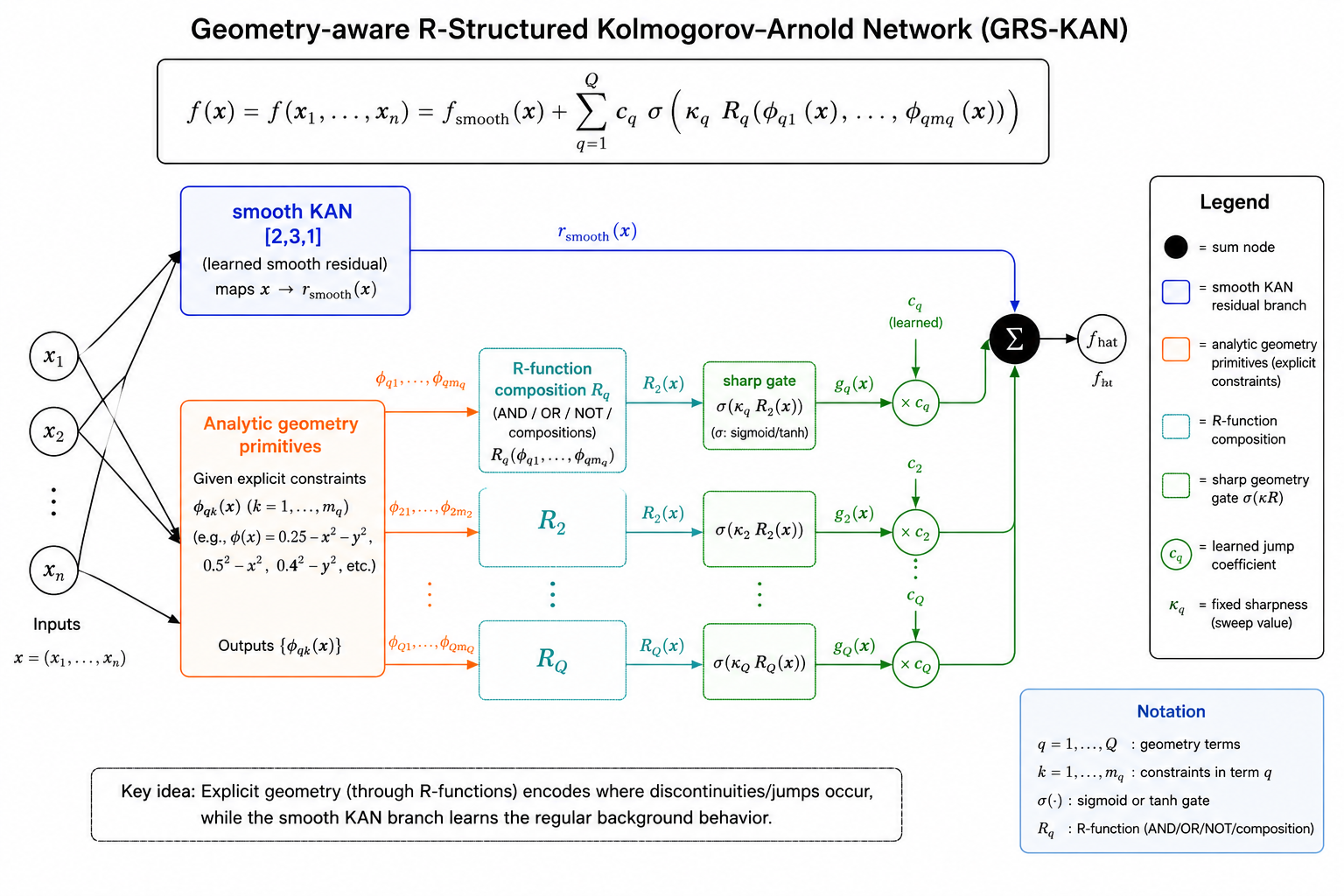}
      \caption{GRS-KAN with geometric gates.}
    \end{subfigure}

  \end{minipage}

  \caption{Comparison of standard KAN and Geometry-aware R-Structured KAN (GRS-KAN). The GRS-KAN augments the base KAN with explicit geometric constraints encoded via R-functions, enabling hard-constrained regression and interpretable structure discovery.}
  \label{fig:kan_vs_grskan}
\end{figure}

Figure~\ref{fig:kan_vs_grskan} illustrates the conceptual difference between standard KANs and the proposed Geometry-aware R-Structured KAN (GRS-KAN). In a standard KAN, all geometric structure must be inferred implicitly from the data through spline edge activations. In contrast, the GRS-KAN explicitly incorporates geometric priors through differentiable R-function gates, allowing known region constraints and discontinuities to be encoded analytically.

Although the architecture defined by
Equation~(\ref{eq:kan_layer}) is capable of approximating highly
nonlinear functions, all geometric information---including feasible
regions, implicit boundaries, Boolean constraints, and
discontinuities---must be inferred implicitly from the training data.
For many scientific and engineering applications, however, such
geometric information is available \emph{a priori}. The next section
introduces R-functions, which provide an analytical framework for
encoding this geometry explicitly. In Section~3 these analytical
geometric representations are combined with the smooth approximation
capability of KANs to form the proposed Geometry-aware R-Structured
Kolmogorov--Arnold Network (GRS-KAN).

\subsection{R-Functions and Geometry Primitives}

R-functions are real-valued functions whose signs are completely determined by the signs of their arguments, enabling logical operations on regions to be represented analytically~\cite{rvachev1982,rvachev1963,shapiro1991,shapiro2007}. In the convention used here, an implicit function $\phi(\mathbf{x}) \geq 0$ denotes the interior of a region.

The analytical representation of implicit geometric regions using
R-functions has recently been developed by Kucherenko \emph{et al.}
\cite{kucherenko2025rfunctions,kucherenko2025designspace}
for the identification of feasible regions and process design spaces in
chemical engineering. The present work builds upon these developments by
embedding analytical R-function representations directly into the KAN
framework, thereby enabling geometry-aware neural architectures.

For two implicit functions $\phi_1(\mathbf{x})$ and $\phi_2(\mathbf{x})$, the R-conjunction (AND) and R-disjunction (OR) are defined as:

\begin{equation}
\phi_1 \wedge_{\alpha} \phi_2 = \frac{\phi_1 + \phi_2 - \sqrt{\phi_1^2 + \phi_2^2 - 2\alpha \phi_1 \phi_2}}{1 + \alpha}, \qquad -1 < \alpha \leq 1,
\label{eq:rand}
\end{equation}

\begin{equation}
\phi_1 \vee_{\alpha} \phi_2 = \frac{\phi_1 + \phi_2 + \sqrt{\phi_1^2 + \phi_2^2 - 2\alpha \phi_1 \phi_2}}{1 + \alpha}, \qquad -1 < \alpha \leq 1,
\label{eq:ror}
\end{equation}
where $\alpha$ controls the smoothness of the operation. When $\alpha = 1$, these reduce to the standard min/max functions:
\[
\phi_1 \wedge_{1} \phi_2 = \min(\phi_1, \phi_2), \qquad \phi_1 \vee_{1} \phi_2 = \max(\phi_1, \phi_2).
\]
When $\alpha = 0$, they become differentiable functions based on the Euclidean norm:
\[
\phi_1 \wedge_{0} \phi_2 = \phi_1 + \phi_2 - \sqrt{\phi_1^2 + \phi_2^2}, \qquad \phi_1 \vee_{0} \phi_2 = \phi_1 + \phi_2 + \sqrt{\phi_1^2 + \phi_2^2}.
\]
The R-negation (NOT) is simply:
\begin{equation}
\neg \phi_1 = -\phi_1.
\label{eq:rnot}
\end{equation}

These operations satisfy De Morgan's laws and form a sufficiently complete system for constructing any Boolean function as a differentiable real-valued function. This makes them particularly suitable for gradient-based optimization in neural networks, as they provide smooth approximations of logical conditions while preserving exactness at the decision boundary.

A composite geometric region $D \subseteq \mathbb{R}^n$ can be described as:
\begin{equation}
D = F[(\phi_1 \geq 0), \ldots, (\phi_m \geq 0)],
\label{eq:composite_region}
\end{equation}
where $\phi_i(\mathbf{x}) \geq 0$ define primitive geometric regions, and $F$ is a set function constructed using standard set operations $\cap$, $\cup$, $\setminus$ on these primitive regions. The function $F$ can be viewed as a Boolean function, in which the set operations are replaced with the corresponding logical functions $\wedge$, $\vee$, $\neg$. We seek a single real-function inequality:
\begin{equation}
R(\phi_1, \ldots, \phi_m) \geq 0
\label{eq:r_function_inequality}
\end{equation}
that defines the composite object $D$ as a closed subset of $\mathbb{R}^n$.

\subsection{Explicit Analytical Construction of a Rectangle Indicator}

For the rectangular tests in this paper, we set $\alpha = 0$, yielding the simplified R-conjunction:
\begin{equation}
\phi_1 \wedge_0 \phi_2 = \phi_1 + \phi_2 - \sqrt{\phi_1^2 + \phi_2^2}.
\label{eq:rand0}
\end{equation}

The rectangle used in the experiments is defined by two perpendicular strips:
\begin{equation}
\phi_1(x,y) = 0.5^2 - x^2, \qquad \phi_2(x,y) = 0.4^2 - y^2,
\label{eq:rect_strips}
\end{equation}
where $\phi_1(x,y) \geq 0$ represents the vertical strip $|x| \leq 0.5$ and $\phi_2(x,y) \geq 0$ represents the horizontal strip $|y| \leq 0.4$. Their R-conjunction yields the rectangle interior:
\begin{equation}
\Rrect(x,y) = \phi_1 \wedge_0 \phi_2.
\label{eq:rect_rfunction}
\end{equation}
Using the above definitions, the expression can be written explicitly as:
\begin{equation}
\Rrect(x,y) = \bigl(0.25 - x^2\bigr) + \bigl(0.16 - y^2\bigr) - \sqrt{\bigl(0.25 - x^2\bigr)^2 + \bigl(0.16 - y^2\bigr)^2}.
\label{eq:rect_explicit}
\end{equation}

This single smooth function completely characterizes the rectangular region:
\begin{itemize}
  \item $\Rrect(x,y) > 0$ for points strictly inside the rectangle $(|x|<0.5,\ |y|<0.4)$,
  \item $\Rrect(x,y) = 0$ on the boundary $(|x|=0.5\ \text{or}\ |y|=0.4)$,
  \item $\Rrect(x,y) < 0$ for points outside the rectangle.
\end{itemize}

The expression is continuously differentiable everywhere except at the four corner points $(\pm0.5,\ \pm0.4)$, where the gradient is discontinuous. This is unavoidable for a sharp rectangle, but the R-function formulation concentrates the non-smoothness at the corners rather than along entire edges. The gradient for interior points and edge points (excluding corners) is:

\begin{align}
\frac{\partial \Rrect}{\partial x} &= -2x \left(1 - \frac{0.25 - x^2}{\sqrt{(0.25 - x^2)^2 + (0.16 - y^2)^2}}\right), \label{eq:grad_x} \\
\frac{\partial \Rrect}{\partial y} &= -2y \left(1 - \frac{0.16 - y^2}{\sqrt{(0.25 - x^2)^2 + (0.16 - y^2)^2}}\right). \label{eq:grad_y}
\end{align}

The function behaves like a softened indicator that can be sharpened by applying a sigmoid with large $\kappa$:
\begin{equation}
g(x,y) = \sigma\bigl(\kappa\,\Rrect(x,y)\bigr) = \frac{1}{1+\exp\bigl(-\kappa\,\Rrect(x,y)\bigr)},
\label{eq:sigmoid_gate}
\end{equation}
which approaches the ideal indicator $I(|x|<0.5,\ |y|<0.4)$ as $\kappa \to \infty$.

The rectangle is used throughout this paper as the simplest analytical geometric primitive for demonstrating the proposed framework. For such simple regions, equivalent smooth gates can also be constructed directly from products of sigmoid functions or other differentiable indicator approximations. The advantage of the proposed R-function formulation is not the smoothing itself, but its ability to represent arbitrary implicit regions and Boolean combinations of constraints analytically within a unified differentiable framework.

\section{GRS-KAN Architecture}

We propose three variants of the Geometry-aware R-Structured KAN, each suited to different types of geometric priors. 
The three GRS-KAN architectures presented below correspond to increasingly general assumptions 
regarding the interaction between a smooth function and known geometric information: additive, multiplicative, and automatically discovered.

\subsection{Targeted Additive GRS-KAN}

Many scientific and engineering problems involve target functions that
consist of a smooth background together with a localized additive
correction acting only inside a prescribed region. Such functions can be
written in the general form

\begin{equation}
f(x)=f_{\mathrm{smooth}}(x)+c\,I(x\in D),
\label{eq:additive_general}
\end{equation}
where $f_{\mathrm{smooth}}$ is a continuous background function,
$D\subset\mathbb{R}^n$ denotes a known geometric region, $I(\cdot)$ is the
indicator function, and $c$ is the magnitude of the additive jump.
Representative examples include localized source terms, regime changes,
activation phenomena, and discontinuities whose support is known
\emph{a priori}.

The discontinuous indicator function is unsuitable for gradient-based
optimization. In the proposed GRS-KAN framework, it is replaced by a
differentiable approximation constructed from an analytical R-function.
Specifically, the indicator is approximated by

\[
I(x\in D)\approx \sigma(\kappa R(x)),
\]
where $R(x)$ is the R-function representation of the region $D$,
$\sigma(\cdot)$ is the logistic sigmoid, and $\kappa$ controls the
transition sharpness.

For the rectangular benchmark considered in this paper,
$R(x)$ is given by the analytical rectangle R-function
$R_{\mathrm{rect}}(x,y)$ defined in Section~2.3. The targeted additive
GRS-KAN therefore becomes

\begin{equation}
\hat f(x,y)
=
f_{\mathrm{smooth}}(x,y)
+
c\,\sigma\!\left(\kappa R_{\mathrm{rect}}(x,y)\right),
\label{eq:additive_grskan}
\end{equation}
where $f_{\mathrm{smooth}}$ is represented by a standard KAN branch that
learns the continuous background, while the second term explicitly models
the localized discontinuity through the analytical geometry. The jump
amplitude $c$ is learned during training, whereas $\kappa$ controls the
sharpness of the transition across the boundary.

\subsection{Targeted Multiplicative GRS-KAN}

A second important class of problems consists of functions whose values
are non-zero only inside a prescribed geometric region. Such functions
can be expressed in the general form

\begin{equation}
f(x)=f_{\mathrm{smooth}}(x)\,I(x\in D),
\label{eq:multiplicative_general}
\end{equation}
where $f_{\mathrm{smooth}}$ denotes the underlying continuous function,
$D\subset\mathbb{R}^n$ is a known geometric support, and $I(\cdot)$ is the
indicator function. This formulation naturally arises when the geometry
determines the support, visibility, activation, or amplitude of the
underlying physical process.

As in the additive formulation, the discontinuous indicator function is
replaced by a differentiable approximation based on an analytical
R-function,

\[
I(x\in D)\approx \sigma(\kappa R(x)),
\]
where $R(x)$ is the R-function representation of the support region,
$\sigma(\cdot)$ is the logistic sigmoid, and $\kappa$ controls the
sharpness of the transition.

For the rectangular benchmark considered in this paper,
$R(x)$ is represented by the analytical rectangle R-function
$R_{\mathrm{rect}}(x,y)$ introduced in Section~2.3. The targeted
multiplicative GRS-KAN is therefore defined as

\begin{equation}
\hat f(x,y)
=
f_{\mathrm{smooth}}(x,y)\,
c\,\sigma\!\left(\kappa R_{\mathrm{rect}}(x,y)\right),
\label{eq:multiplicative_grskan}
\end{equation}
where the KAN branch learns the smooth background function, while the
R-function gate explicitly encodes the geometric support. The trainable
coefficient $c$ scales the response, and $\kappa$ controls the sharpness
of the transition near the boundary.

Typical examples include masked functions, visibility constraints,
domain-restricted solutions of partial differential equations,
region-dependent material properties, and financial payoffs activated
only inside prescribed state-space regions. The benchmark considered in
this paper corresponds to

\[
f(x,y)=xy\,I\!\left((x,y)\in D\right),
\]
where $D$ is the rectangular region represented analytically by
$R_{\mathrm{rect}}(x,y)$.

\subsection{Agnostic GRS-KAN with Learnable Structure-Selection Parameters}

In many practical applications, the manner in which geometric information
interacts with the smooth background is not known \emph{a priori}. The
target function may be purely smooth, contain an additive localized
correction, exhibit multiplicative geometric support, or involve a
combination of these mechanisms. Rather than selecting a specific
architecture beforehand, we introduce an agnostic GRS-KAN that
simultaneously considers all three possibilities and learns their relative
importance directly from the data.

The agnostic architecture combines a baseline KAN branch with additive
and multiplicative geometry-aware branches,

\begin{equation}
\hat f(x,y)
=
w_{\mathrm{KAN}}\,f_{\mathrm{base}}(x,y)
+
w_{\mathrm{add}}\,g(x,y)
+
w_{\mathrm{mul}}\,f_{\mathrm{mask}}(x,y)\,g(x,y),
\label{eq:agnostic_grskan}
\end{equation}
where
\[
g(x,y)
=
\sigma\!\left(\kappa R_{\mathrm{rect}}(x,y)\right)
\]
is the differentiable R-function gate introduced in the previous
subsections.

The learnable coefficients
$w_{\mathrm{KAN}}$,
$w_{\mathrm{add}}$,
and
$w_{\mathrm{mul}}$
determine the relative contribution of the three branches during
training. Consequently, the proposed architecture performs not only
function approximation but also automatic structure identification.
The coefficients $w_{\mathrm{KAN}}$, $w_{\mathrm{add}}$, and
$w_{\mathrm{mul}}$ act as \emph{learnable structure-selection
parameters}, automatically identifying the most appropriate interaction
between the smooth approximation and the analytical geometric prior.
Depending on the underlying problem, the network may recover a purely
smooth KAN model, a targeted additive representation, a targeted
multiplicative representation, or a hybrid combination of these
components. The learned structure-selection parameters therefore provide
both an interpretable decomposition of the solution and insight into the
underlying functional relationship between geometry and the smooth
background.

\begin{figure}[H]
  \centering
  \begin{subfigure}{1.1\linewidth}
    \centering
    \includegraphics[width=\linewidth]{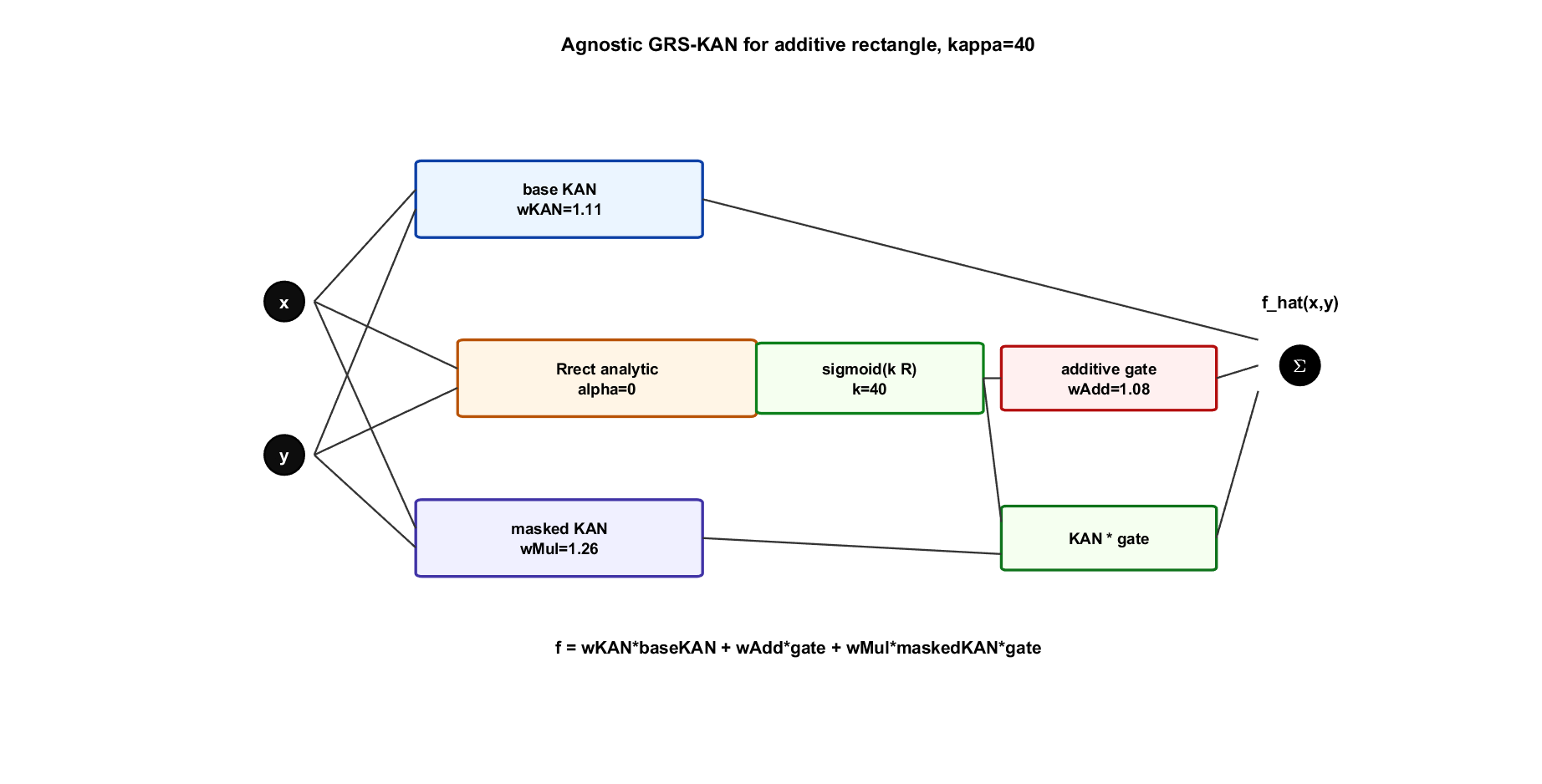}
    \caption{Additive discontinuity problem. All three branches remain active,
    indicating that the optimal approximation combines smooth,
    additive, and multiplicative components.}
    \label{fig:agnostic-additive}
  \end{subfigure}
  \hfill
  \begin{subfigure}{1.1\linewidth}
    \centering
    \includegraphics[width=\linewidth]{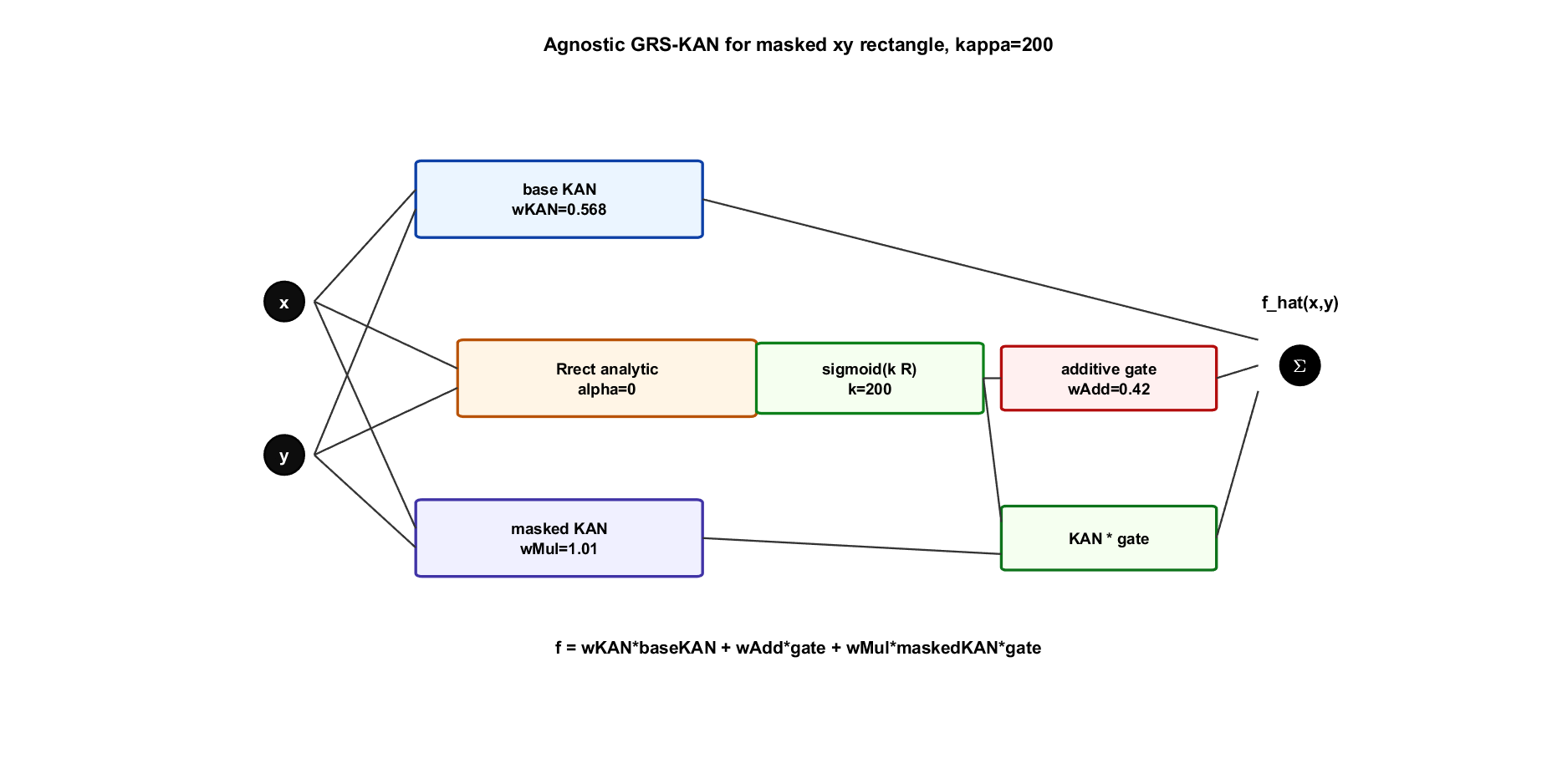}
    \caption{Masked $xy$ problem. The multiplicative geometry-aware branch
    dominates, correctly identifying that the target function is governed
    primarily by geometric support.}
    \label{fig:agnostic-masked}
  \end{subfigure}
  \caption{Agnostic GRS-KAN architecture illustrating the learned
  structure-selection parameters for two benchmark problems. The relative
  branch activations reveal how analytical geometry interacts with the
  smooth KAN approximation.}
  \label{fig:agnostic-architectures}
\end{figure}

Figure~\ref{fig:agnostic-architectures} illustrates that the learned
structure-selection parameters provide a useful diagnostic of the
underlying functional structure. Rather than simply weighting individual
network branches, they reveal how analytical geometry should interact
with the smooth KAN approximation to represent the target function.

For the additive rectangle problem
(Figure~\ref{fig:agnostic-additive}), the KAN, additive, and
multiplicative branches all contribute to the final prediction,
indicating that the agnostic model learns a mixed representation rather
than selecting a single interaction mechanism. In contrast, for the
masked $xy$ problem (Figure~\ref{fig:agnostic-masked}), the
multiplicative branch receives the dominant weight, correctly
identifying that the target function is primarily governed by geometric
support.

These results demonstrate that the proposed agnostic GRS-KAN can
simultaneously perform function approximation and automatic structure
discovery, eliminating the need to prescribe \emph{a priori} how
analytical geometry should interact with the smooth approximation. This
capability makes GRS-KAN applicable to a broad class of scientific
and engineering problems where the functional role of known geometric
constraints is itself unknown.

\section{Numerical Experiments}

All experiments used reproducible random seeds, $N_{\mathrm{train}} = 1000$, and $N_{\mathrm{test}} = 1000$. The primary metrics are train RMSE, test RMSE, parameter count, and boundary-band RMSE (for problems with discontinuities).

\subsection{Validation of Standard KAN on the Liu et al. Toy Benchmark 1}

We first validated our MATLAB KAN implementation against the toy example from Liu et al.~\cite{liu2025kan}:

\begin{equation}
f(x,y)=\exp(\sin(\pi x)+y^2).
\label{eq:toy_example}
\end{equation}

The workflow starts from an improved full $[2,5,1]$ KAN, uses edge-activation scores to identify a compact $[2,1,1]$ structure, retrains/fine-tunes the compact KAN, and finally applies symbolic snapping to recover the underlying analytical representation.

The same data split was also tested using MATLAB's built-in \texttt{fitrnet} regression neural network. Hidden-layer configurations $[20,20]$, $[50,50]$, $[100,100]$, $[200,100]$, and $[100,100,50]$ were swept using ReLU and tanh activations. The selected configuration the lowest validation RMSE was a tanh MLP with hidden-layer sizes $[100,100,50]$, standardization enabled, ridge parameter $\lambda=10^{-6}$, and iteration limit 1200.

\begin{table}[H]
\centering
\caption{Toy problem results for $f(x,y)=\exp(\sin(\pi x)+y^2)$.}
\begin{tabular}{lrrrr}
\toprule
Model & Train RMSE & Test RMSE & Params & Time (s) \\
\midrule
Full KAN $[2,5,1]$ & 0.00355 & 0.00421 & 195 & 11.6 \\
Pruned-structure compact KAN $[2,1,1]$ & 0.00247 & 0.00277 & 51 & 0.540 \\
MATLAB NN, \texttt{fitrnet} $[100,100,50]$/tanh & 0.00229 & 0.00439 & 15501 & 8.02 \\
Symbolic snapped KAN & 0 & 0 & 12 & 0.0500 \\
\bottomrule
\end{tabular}
\label{tab:toy-kan}
\end{table}

\begin{figure}[H]
\hspace*{-0.1\linewidth}
\begin{minipage}{1.2\linewidth}

\begin{subfigure}{\linewidth}
\centering
\includegraphics[width=\linewidth]{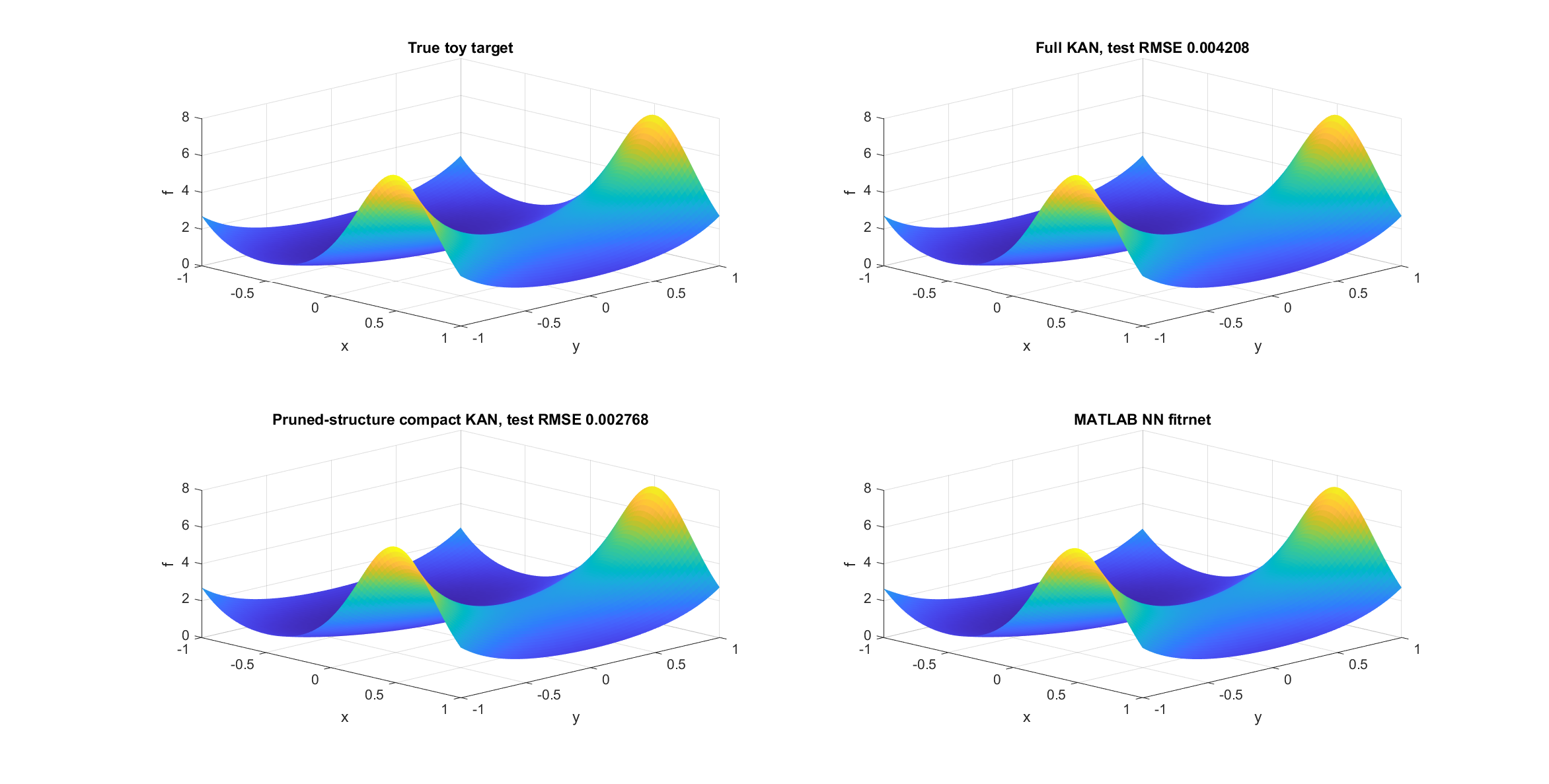}
\caption{True target, full KAN, compact KAN, and MATLAB NN surfaces.}
\end{subfigure}

\vspace{1ex}

\begin{subfigure}{0.8\linewidth}
\centering
\includegraphics[width=\linewidth]{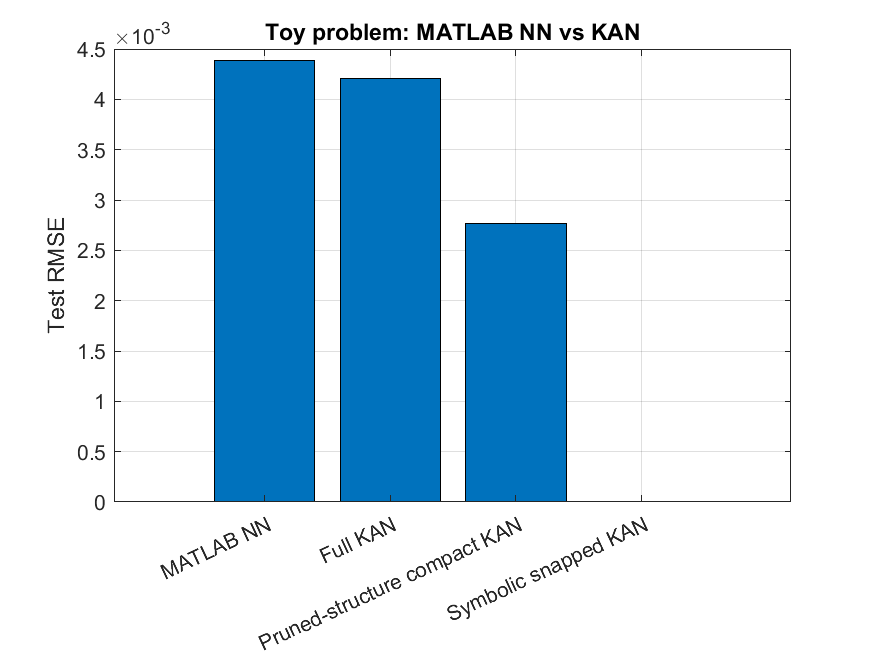}
\caption{Test RMSE comparison for MATLAB NN and KAN variants.}
\end{subfigure}

\end{minipage}

\caption{Toy benchmark comparison including the MATLAB NN baseline.}
\label{fig:toy-nn}
\end{figure}

\begin{figure}[H]
\centering
\begin{subfigure}{0.8\linewidth}
\centering
\includegraphics[width=\linewidth]{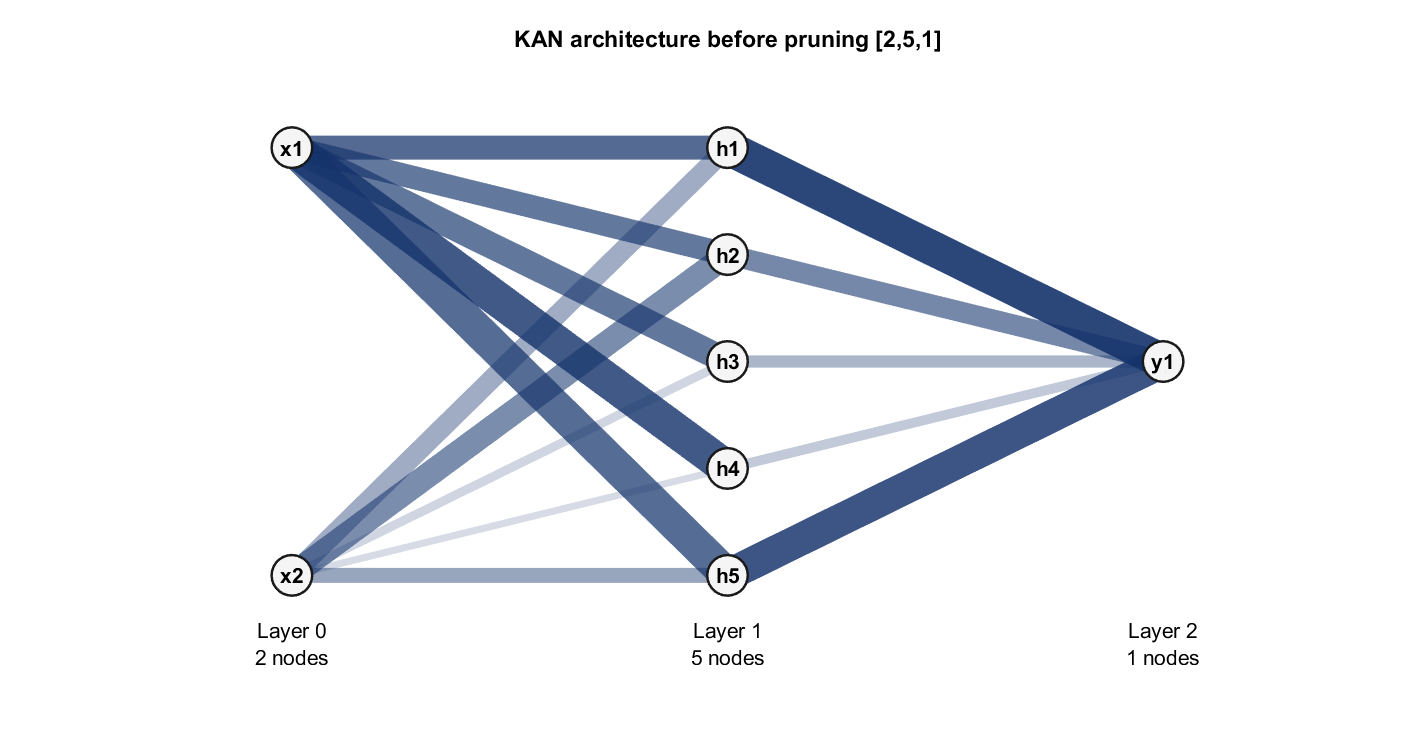}
\caption{Full $[2,5,1]$ KAN before pruning.}
\end{subfigure}
\hfill
\begin{subfigure}{0.8\linewidth}
\centering
\includegraphics[width=\linewidth]{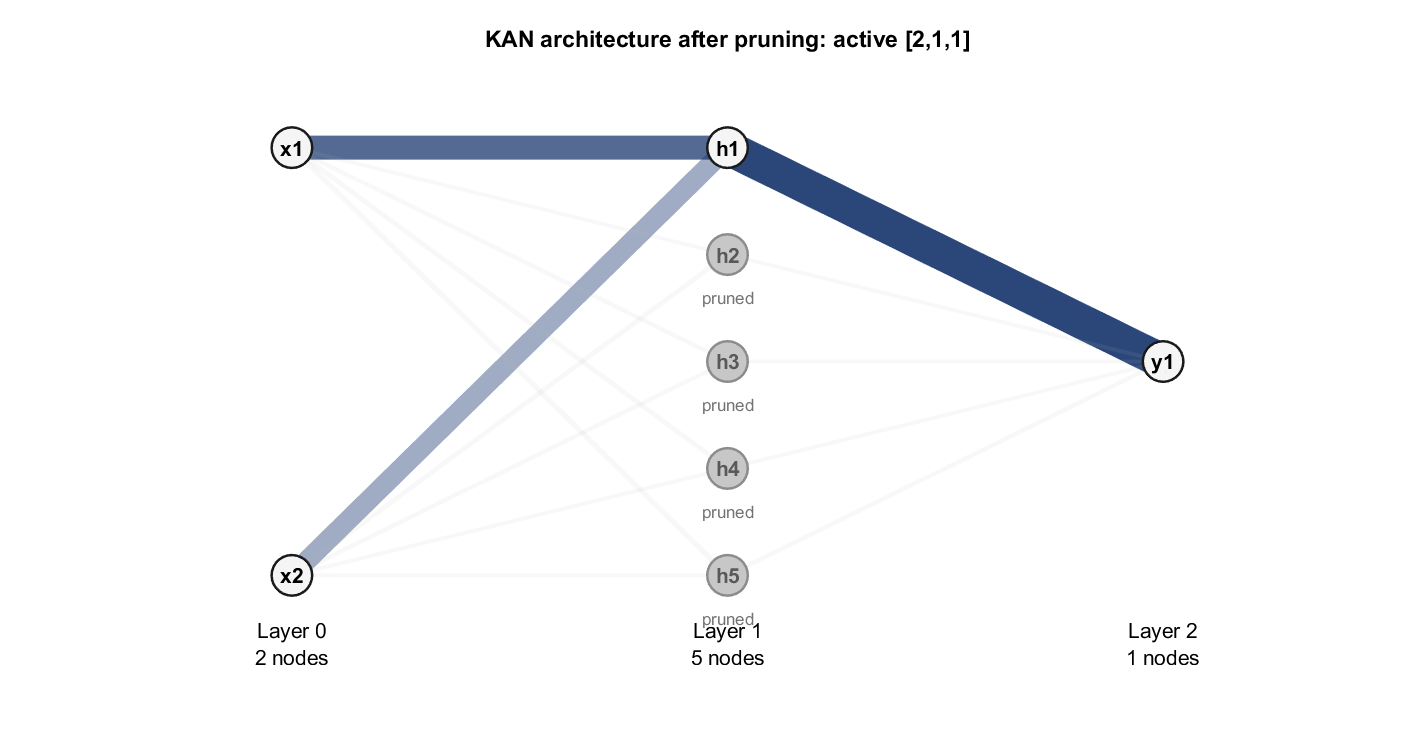}
\caption{Same trained network with inactive hidden nodes shown as pruned.}
\end{subfigure}
\caption{Standard KAN architecture plots. Edge opacity and thickness indicate mean absolute edge activation magnitude. Toy Benchmark 1}
\label{fig:kan-architecture}
\end{figure}

\begin{figure}[H]
\centering
\begin{subfigure}{0.8\linewidth}
\centering
\includegraphics[width=\linewidth]{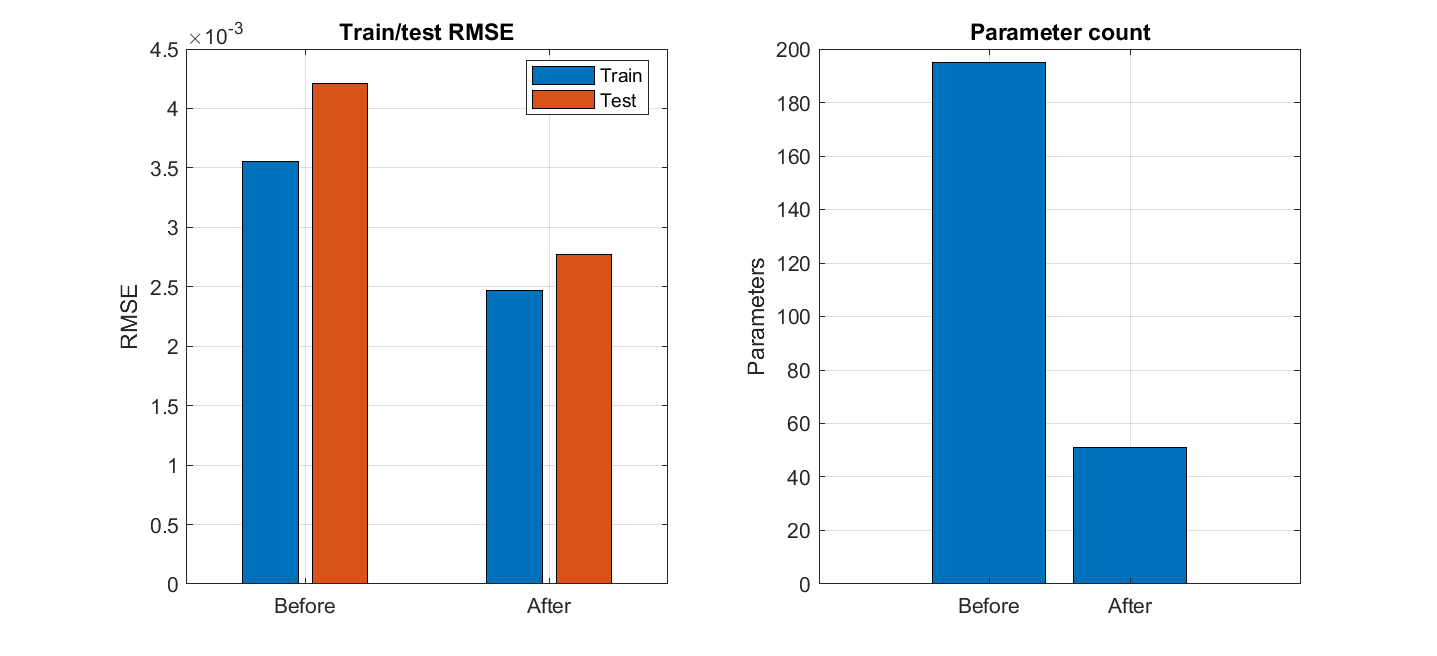}
\caption{Before/after RMSE and parameter count after pruning.}
\end{subfigure}
\hfill
\begin{subfigure}{0.8\linewidth}
\centering
\includegraphics[width=\linewidth]{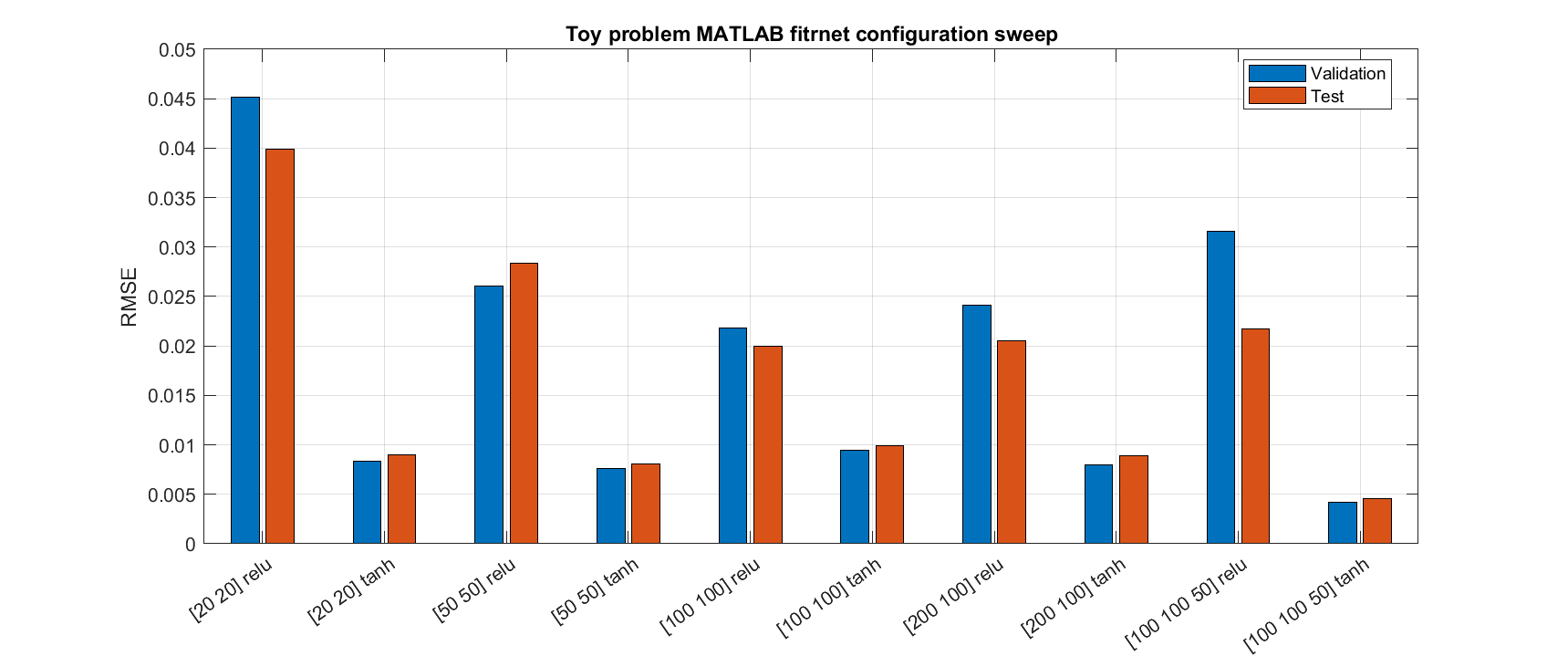}
\caption{MATLAB NN validation/test RMSE over layer sizes and activations.}
\end{subfigure}
\caption{Toy benchmark pruning and NN configuration diagnostics.}
\label{fig:toy-kan-diagnostics}
\end{figure}

Figure~\ref{fig:toy-nn} compares the fitted surfaces and final test errors. All learned models reproduce the smooth nonlinear target accurately. The improved full KAN reaches test RMSE $0.00421$ using only 195 parameters, slightly outperforming the selected MATLAB neural network baseline with test RMSE $0.00439$ despite the latter requiring 15501 parameters. After pruning and retraining, the compact $[2,1,1]$ KAN further improves the test RMSE to $0.00277$ while reducing the parameter count to only 51. The symbolic snapped KAN exactly recovers the analytical expression of the target function.

Figure~\ref{fig:kan-architecture} illustrates the structural pruning behaviour of the trained KAN. Subfigure~(a) shows the full trained $[2,5,1]$ architecture before simplification. Edge thickness and opacity are proportional to the mean absolute edge activation magnitude, so weak functional connections appear pale and thin. Subfigure~(b) shows the resulting pruning decision, where inactive hidden nodes are removed automatically, leaving the compact $[2,1,1]$ architecture.

This behaviour confirms one of the central claims of Liu et al.~\cite{liu2025kan}: KANs can reveal sparse compositional structure rather than acting purely as black-box approximators. In many scientific problems, the effective functional structure is substantially simpler than the original dense architecture, and KAN pruning exposes this reduced representation explicitly.

Figure~\ref{fig:toy-kan-diagnostics} provides additional diagnostics behind this comparison. Subfigure~(a) shows the effect of compacting the KAN structure: the number of parameters decreases from 195 to 51 while the RMSE improves after fine tuning. Thus pruning is not merely compressing the model; it removes weak hidden paths that are not required by the underlying compositional structure.

Subfigure~(b) presents the MATLAB NN configuration sweep. For this smooth globally nonlinear target, tanh activations outperform ReLU activations, and the best validation score is achieved by the deeper $[100,100,50]$ tanh network. The comparison highlights that KANs can achieve competitive or better accuracy using substantially fewer parameters, while retaining the additional capability of symbolic structure recovery.

\subsection{Unconstrained product function}

We next evaluated our KAN implementation on a second benchmark problem from Liu et al.~\cite{liu2025kan}, the unconstrained product function:

\begin{equation}
f(x,y) = xy.
\label{eq:product}
\end{equation}

Unlike the smooth nonlinear target in Benchmark 1, this function has an exact multiplicative structure with no additive components or higher-order nonlinearities. It therefore provides a stringent test of whether the pruning and symbolic snapping procedures can correctly identify the absence of unnecessary terms and recover a purely multiplicative relationship.

Applying the same workflow as before, we initialized a full $[2,5,1]$ KAN, then used edge-activation scores to prune weak connections. The resulting compact architecture was $[2,2,1]$, which retains a single hidden layer with two active nodes. After retraining the pruned structure, we performed symbolic snapping to map the learned edge functions onto a closed-form analytical expression.

As shown in Table~\ref{tab:xy-kan}, pruning reduced the parameter count from 255 to 102 while simultaneously improving both train and test RMSE by more than an order of magnitude. Symbolic snapping then reduced the model to just 9 parameters and recovered the exact product $f(x,y)=xy$ to within machine precision, with test RMSE on the order of $10^{-17}$. This demonstrates that the KAN framework can distinguish true multiplicative structure from spurious correlations induced by overparameterization.

\begin{table}[H]
  \centering
  \caption{KAN results for $f(x,y)=xy$.}
  \begin{tabular}{lrrrr}
    \toprule
    Model & Train RMSE & Test RMSE & Params & Time (s) \\
    \midrule
    Full KAN $[2,5,1]$ & 0.00690 & 0.00720 & 255 & 15.8 \\
    Pruned KAN $[2,2,1]$ & $4.81\!\times 10^{-4}$ & $5.16\!\times 10^{-4}$ & 102 & 0.990 \\
    Symbolic snapped KAN & $2.82\!\times 10^{-17}$ & $2.63\!\times 10^{-17}$ & 9 & 0.0500 \\
    \bottomrule
  \end{tabular}
  \label{tab:xy-kan}
\end{table}

\begin{figure}[H]
  \centering
  \begin{subfigure}{0.8\linewidth}
    \centering
    \includegraphics[width=\linewidth]{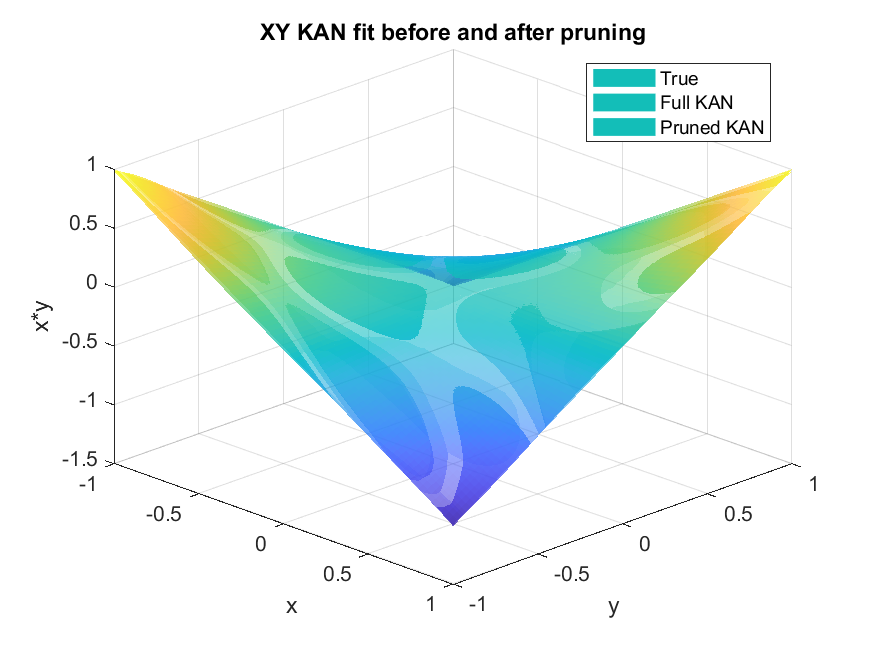}
    \caption{Surface fit for full and pruned KAN.}
  \end{subfigure}
  \hfill
  \begin{subfigure}{0.8\linewidth}
    \centering
    \includegraphics[width=\linewidth]{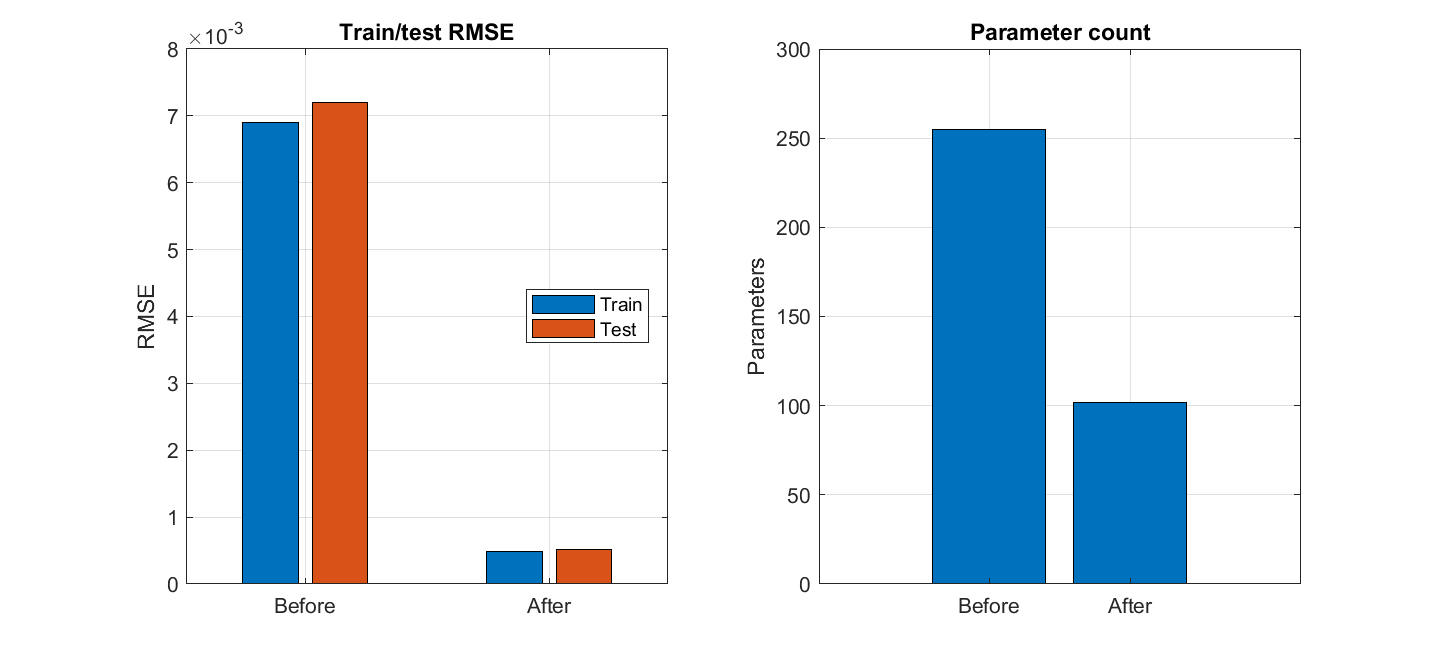}
    \caption{RMSE and parameter count before and after pruning.}
  \end{subfigure}
  \caption{Unconstrained product KAN results.}
  \label{fig:xy-kan}
\end{figure}

Figure~\ref{fig:xy-kan} provides a simple symbol-discovery check. Both the full and
pruned KAN reproduce the saddle-shaped $xy$ surface, while pruning reduces the architecture
from $[2,5,1]$ to $[2,2,1]$ and lowers the test RMSE from $7.20\times10^{-3}$ to
$5.16\times10^{-4}$. This is consistent with the representation
$xy=((x+y)^2-(x-y)^2)/4$, which naturally requires two hidden directions.

These results verify that our MATLAB baseline captures the central KAN ideas: edge activations, summation nodes, spline functions, residual SiLU basis, pruning by edge activation magnitudes, and symbolic simplification.

\subsection{Rectangular Additive Discontinuity}

The first GRS-KAN benchmark introduces a discontinuous jump superimposed on a smooth background:
\begin{equation}
  f(x,y)=\sin(\pi x)+y^2+2I(|x|<0.5,\ |y|<0.4),
  \label{eq:rect-additive-target}
\end{equation}
where the rectangular region is encoded analytically through $R_{\mathrm{rect}}$ and the additive model in Eq.~\eqref{eq:additive_grskan}.

For comparison, we also evaluate a standard multilayer perceptron baseline using MATLAB's \texttt{fitrnet}. A hyperparameter sweep over hidden-layer configurations $[20,20]$, $[50,50]$, $[100,100]$, $[200,100]$, and $[100,100,50]$ with both ReLU and tanh activations was performed using a 20\% validation split. The selected network---a ReLU MLP with $[20,20]$ hidden units, standardization, ridge parameter $\lambda=10^{-6}$, and 1200 iterations---was retrained on the full training set prior to final evaluation.

\begin{table}[H]
  \centering
  \caption{Rectangular additive discontinuity results including the MATLAB NN baseline.}
  \begin{tabular}{lrrrrr}
    \toprule
    Model & $\kappa$ & Train RMSE & Test RMSE & Boundary RMSE & Params \\
    \midrule
    Standard KAN & -- & 0.128 & 0.277 & 0.612 & 255 \\
    MATLAB NN, \texttt{fitrnet} $[20,20]$/ReLU & -- & 0.041 & 0.114 & 0.365 & 501 \\
    Generic RS-KAN & -- & 0.105 & 0.267 & 0.600 & 416 \\
    Additive GRS-KAN & 20 & 0.198 & 0.242 & 0.577 & 154 \\
    Additive GRS-KAN & 40 & 0.156 & 0.183 & 0.454 & 154 \\
    Additive GRS-KAN & 80 & 0.127 & 0.140 & 0.353 & 154 \\
    Additive GRS-KAN & 120 & 0.110 & 0.116 & 0.299 & 154 \\
    Additive GRS-KAN & 200 & 0.090 & 0.092 & 0.239 & 154 \\
    \bottomrule
  \end{tabular}
  \label{tab:rect-additive}
\end{table}

\begin{figure}[H]
  \hspace*{-0.1\linewidth}
  \begin{minipage}{1.2\linewidth}
    \begin{subfigure}{\linewidth}
      \centering
      \includegraphics[width=\linewidth]{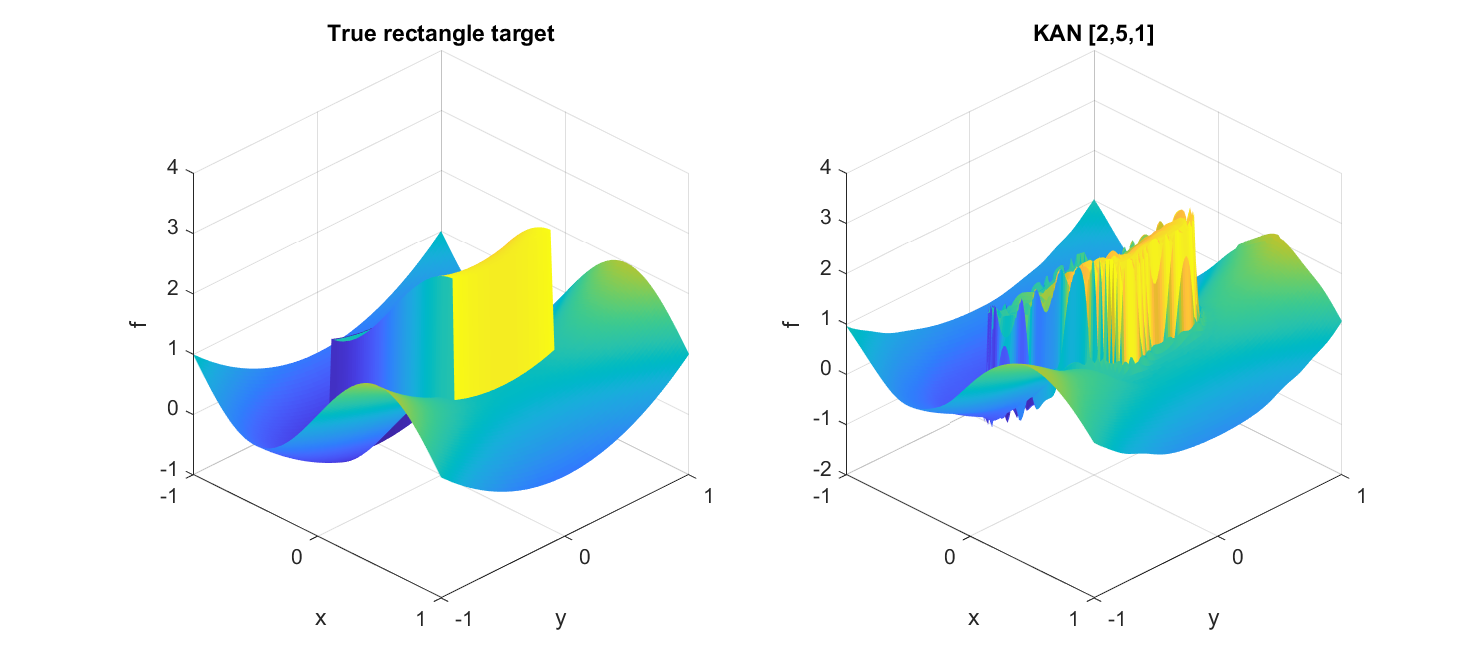}
      \caption{True target (left) and standard KAN $[2,5,1]$ prediction (right).}
    \end{subfigure}

    \vspace{1ex}

    \begin{subfigure}{\linewidth}
      \centering
      \includegraphics[width=\linewidth]{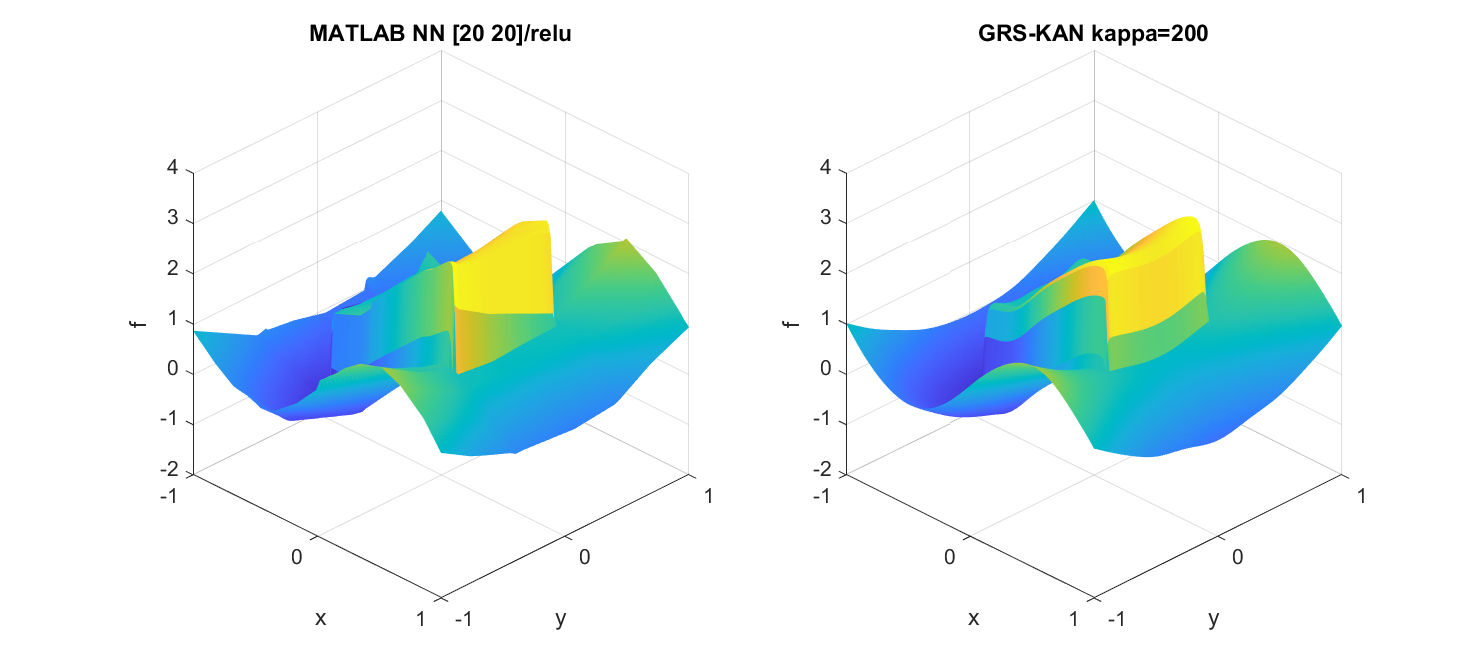}
      \caption{MATLAB NN $[20,20]$/ReLU (left) and GRS-KAN with $\kappa=200$ (right).}
    \end{subfigure}
  \end{minipage}

  \caption{Surface comparisons for the rectangular additive discontinuity.}
  \label{fig:rect-additive-surfaces}
\end{figure}

Figure~\ref{fig:rect-additive-surfaces} compares the fitted surfaces. The standard KAN (subfigure a) produces a visibly smooth transition across the discontinuity, as it must approximate the jump using smooth spline bases. The MATLAB NN (subfigure b) captures the discontinuity more sharply, reflected in its lower test RMSE ($0.1136$ versus $0.2774$). However, the GRS-KAN achieves the sharpest rectangular jump and the best overall test and boundary-band errors, because the jump location is supplied explicitly through $R_{\mathrm{rect}}$ rather than inferred from data alone.

\begin{figure}[H]
  \centering
  \begin{subfigure}{1.0\linewidth}
    \centering
    \includegraphics[width=\linewidth]{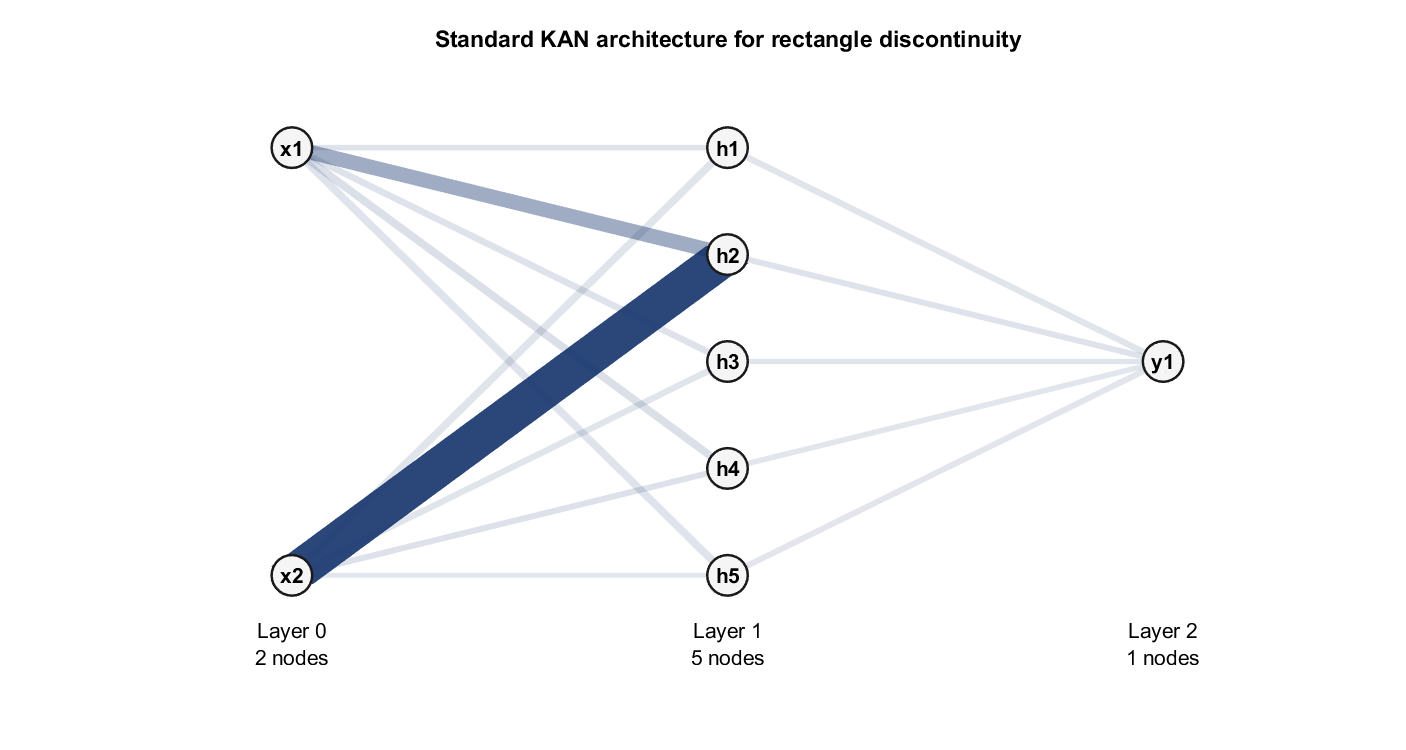}
    \caption{Standard KAN architecture.}
  \end{subfigure}
  \hfill
  \begin{subfigure}{1.2\linewidth}
    \centering
    \includegraphics[width=\linewidth]{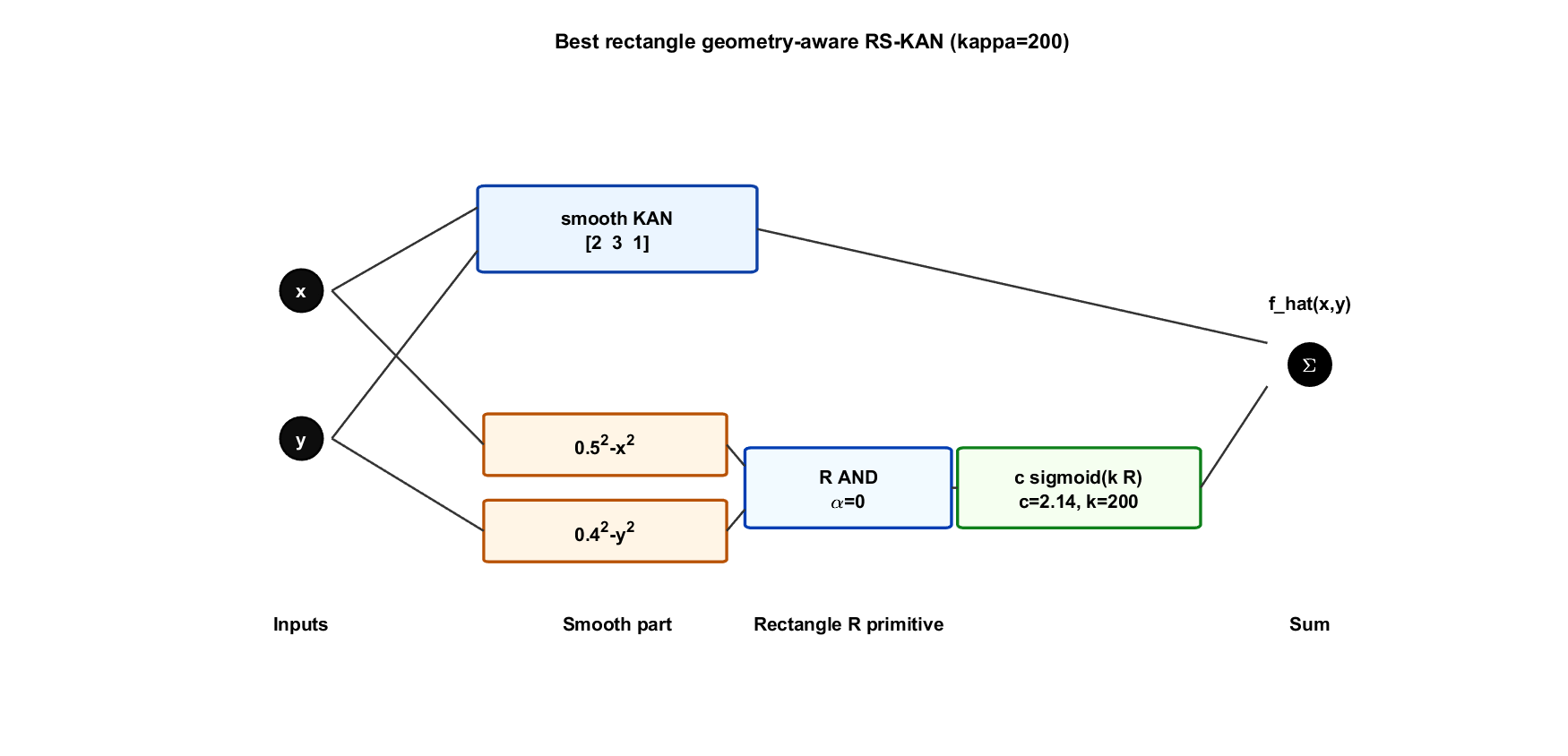}
    \caption{Targeted additive GRS-KAN architecture.}
  \end{subfigure}
  \caption{Architectural comparison for the rectangular discontinuity benchmark.}
  \label{fig:rect-architectures}
\end{figure}

Figure~\ref{fig:rect-architectures} illustrates the architectural difference between the two models. The standard KAN must learn both the smooth background and the discontinuity through its spline edge activations, whereas the targeted GRS-KAN separates these roles: the KAN branch models the smooth residual, while the R-function gate encodes the jump explicitly. This structural decoupling is the primary source of the accuracy improvements reported below.

\begin{figure}[H]
  \hspace*{-0.05\linewidth}
  \begin{minipage}{1.1\linewidth}

    \begin{subfigure}{0.7\linewidth}
      \centering
      \includegraphics[width=\linewidth]{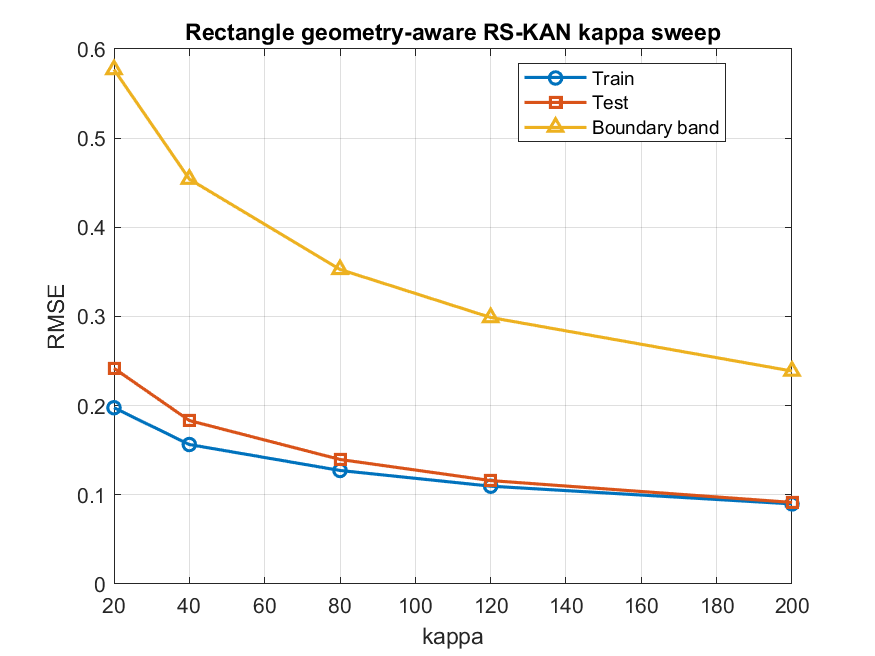}
      \caption{Train, test, and boundary-band RMSE versus $\kappa$.}
      \label{fig:rect-kappa}
    \end{subfigure}

    \vspace{1ex}

    \begin{subfigure}{\linewidth}
      \centering
      \includegraphics[width=\linewidth]{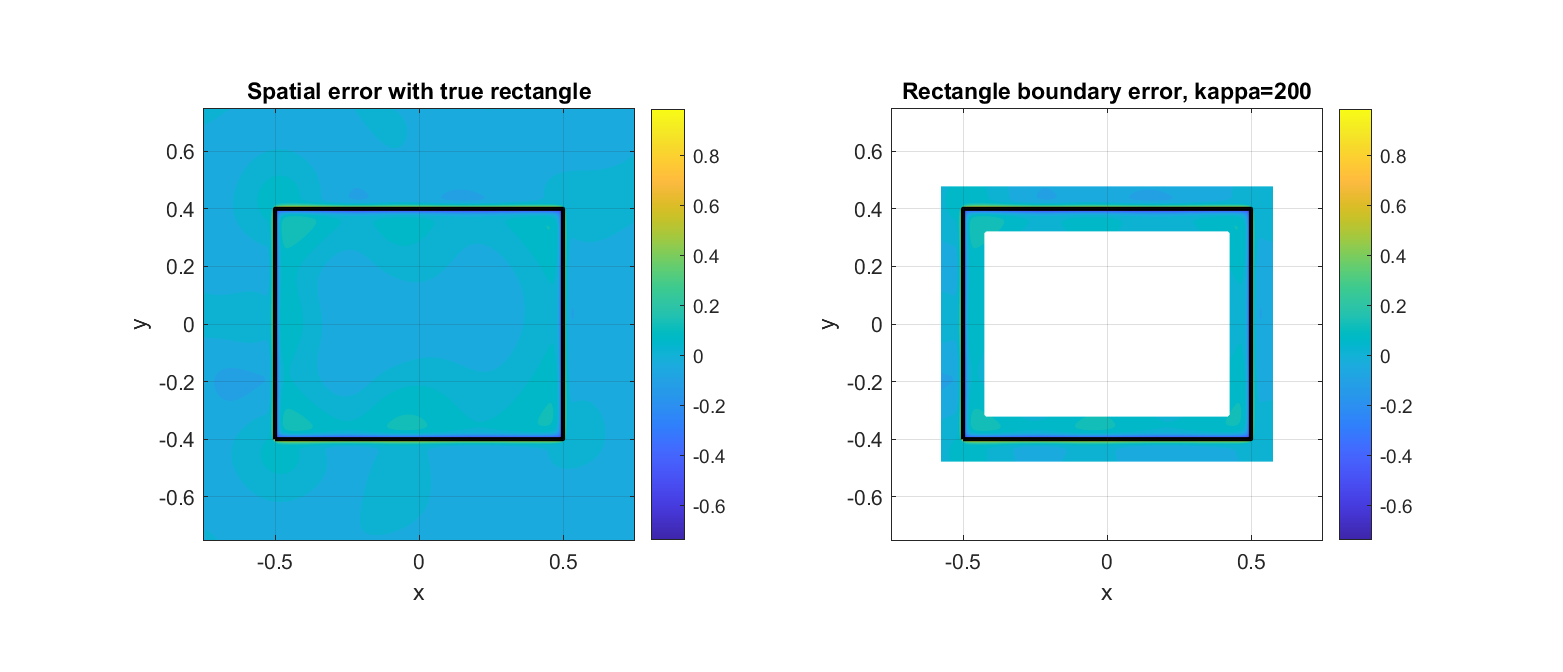}
      \caption{Spatial error map with the true rectangle boundary overlaid.}
      \label{fig:rect-error}
    \end{subfigure}

  \end{minipage}

  \caption{RMSE analysis and spatial error localization.}
  \label{fig:rect-rmse-error}
\end{figure}

Figure~\ref{fig:rect-rmse-error} provides a quantitative and spatial analysis. Subfigure (a) shows that increasing $\kappa$ sharpens the sigmoid gate and monotonically improves both test and boundary-band RMSE over the range tested, with $\kappa=200$ yielding the best performance in this experiment. Subfigure (b) localizes the remaining error primarily near the rectangular boundary, where the true discontinuity is most difficult to approximate with a smooth gate. This concentration of error is expected and indicates that the model correctly identifies the geometric structure.

Quantitatively, the targeted additive GRS-KAN with $\kappa=200$ reduces test RMSE from $0.2774$ (standard KAN) to $0.0915$---a reduction of $67\%$---and improves boundary-band RMSE from $0.6116$ to $0.2388$, a reduction of $61\%$. These gains demonstrate that explicit geometric encoding via R-functions substantially improves both global accuracy and localization of discontinuities.

\subsection{Masked Product with Rectangular Support}

The multiplicative geometry test enforces support only inside the rectangle:

\begin{equation}
f(x,y) = xy \, I(|x|<0.5,\ |y|<0.4).
\label{eq:masked_xy_target}
\end{equation}

\begin{table}[H]
  \centering
  \caption{Masked $xy$ rectangle results.}
  \begin{tabular}{lrrrrr}
    \toprule
    Model & $\kappa$ & Train RMSE & Test RMSE & Boundary RMSE & Params \\
    \midrule
    Standard KAN & -- & 0.00908 & 0.0149 & 0.0302 & 255 \\
    Multiplicative GRS-KAN & 20 & 0.0102 & 0.0134 & 0.0298 & 154 \\
    Multiplicative GRS-KAN & 40 & 0.00769 & 0.0107 & 0.0266 & 154 \\
    Multiplicative GRS-KAN & 80 & 0.00708 & 0.00780 & 0.0218 & 154 \\
    Multiplicative GRS-KAN & 120 & 0.00616 & 0.00675 & 0.0187 & 154 \\
    Multiplicative GRS-KAN & 200 & 0.00405 & 0.00620 & 0.0152 & 154 \\
    \bottomrule
  \end{tabular}
  \label{tab:masked-xy}
\end{table}

\begin{figure}[H]
    \hspace*{-0.2\linewidth}
    \includegraphics[width=1.4\linewidth]{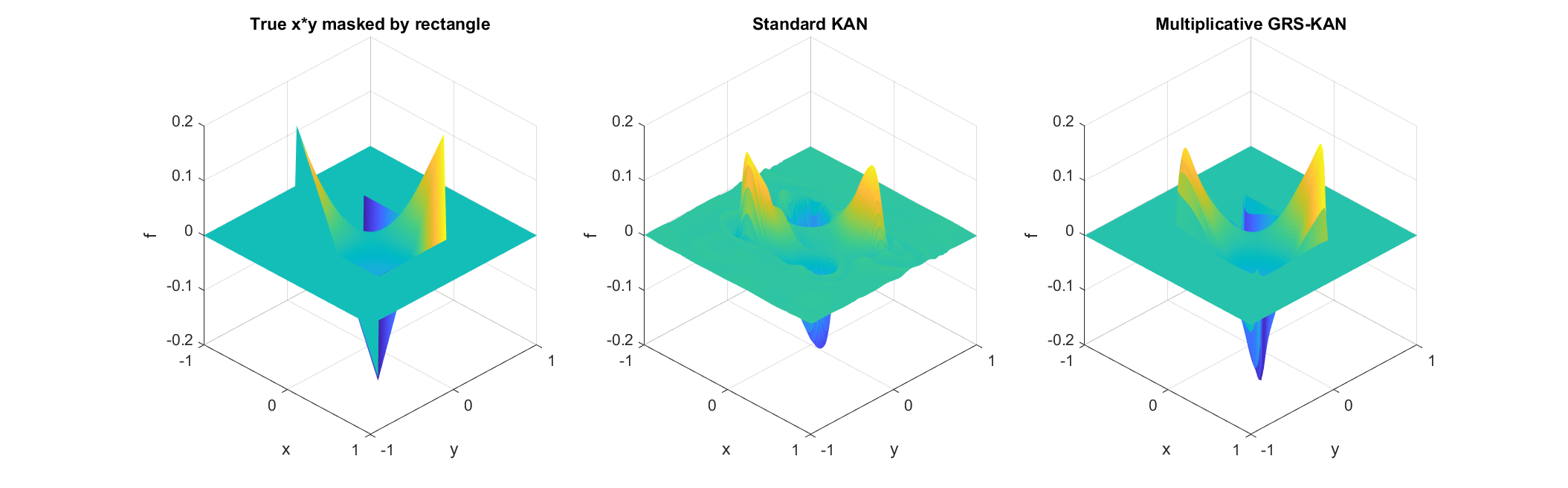}
    \caption{Predicted surfaces for $xy$ masked by the rectangle.}
    \label{fig:masked-xy-fit}
\end{figure}

\begin{figure}[H]
  \centering
  \begin{subfigure}{0.6\linewidth}
    \centering
    \includegraphics[width=\linewidth]{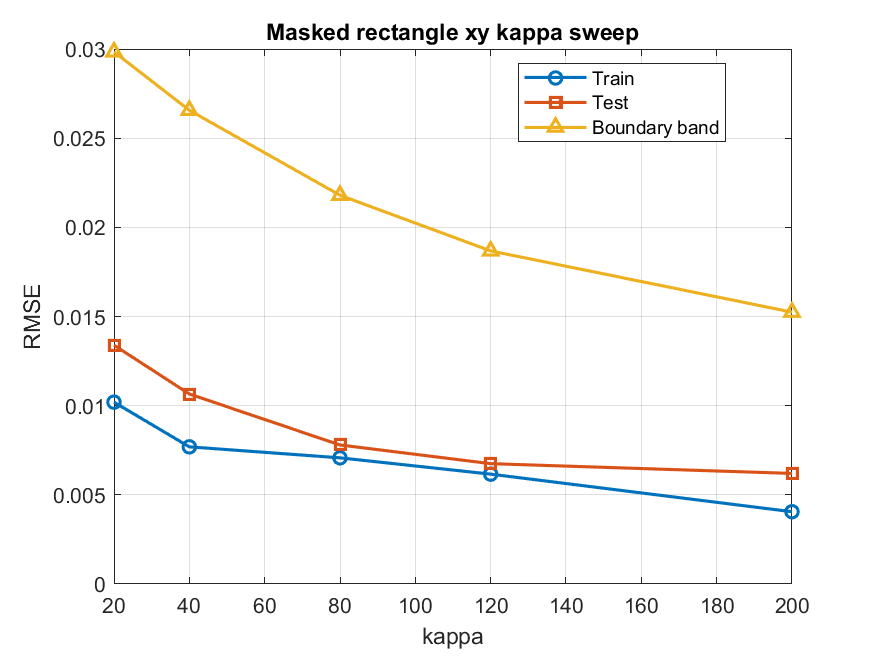}
    \caption{Train, test, and boundary-band RMSE versus $\kappa$.}
    \label{fig:masked-xy-kappa}
  \end{subfigure}
  \hfill
  \begin{subfigure}{0.6\linewidth}
    \centering
    \includegraphics[width=\linewidth]{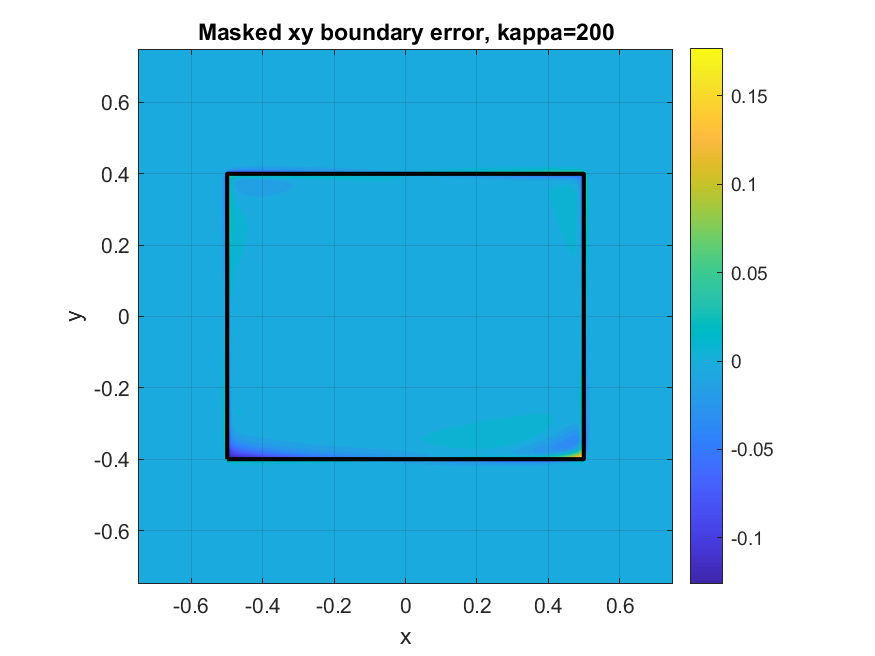}
    \caption{Spatial error map with the true rectangle boundary overlaid.}
    \label{fig:masked-xy-error}
  \end{subfigure}
  \caption{Masked $xy$ rectangle plots: (a) RMSE sweep and (b) error map.}
  \label{fig:masked-xy-subplots}
\end{figure}

Figure~\ref{fig:masked-xy-fit} illustrates the case where the multiplicative GRS-KAN is
structurally appropriate. The target is zero outside the rectangle and behaves like
$xy$ inside it. A conventional KAN must simultaneously learn both the smooth product-like variation and the underlying geometric support from data. The multiplicative GRS-KAN decouples these tasks: the KAN branch models the smooth function, whereas the analytical R-function representation explicitly encodes the geometric support. 
The remaining
errors are concentrated near the rectangle boundary and corners, as shown in 
Figure~\ref{fig:masked-xy-error}, where the smooth sigmoid gate approximates a hard indicator. 
Figure~\ref{fig:masked-xy-kappa} further shows the trade-off in choosing the sharpness 
parameter $\kappa$.

The multiplicative GRS-KAN with $\kappa=200$ reduced test RMSE from $0.0149$ to $0.0062$ ($58\%$ reduction) and boundary RMSE from $0.0302$ to $0.0152$ ($50\%$ reduction).

\subsection{Agnostic GRS-KAN with Learned Branch Weights}

The agnostic model in Eq.~\eqref{eq:agnostic_grskan} was tested on both rectangular targets. The learned weights reveal the preferred structure.

\begin{table}[H]
  \centering
  \caption{Best agnostic GRS-KAN results on rectangular tasks.}
  \begin{tabular}{lrrrrrrr}
    \toprule
    Task & $\kappa$ & Test RMSE & Boundary RMSE & $w_{\mathrm{KAN}}$ & $w_{\mathrm{add}}$ & $w_{\mathrm{mul}}$ & Params \\
    \midrule
    Additive discontinuity & 40 & 0.100 & 0.322 & 1.11 & 1.08 & 1.26 & 309 \\
    Masked $xy$ & 200 & 0.00477 & 0.0146 & 0.568 & 0.420 & 1.01 & 309 \\
    \bottomrule
  \end{tabular}
  \label{tab:agnostic-rect}
\end{table}

\begin{figure}[H]
  \centering
  \begin{subfigure}{1.2\linewidth}
    \centering
    \includegraphics[width=\linewidth]{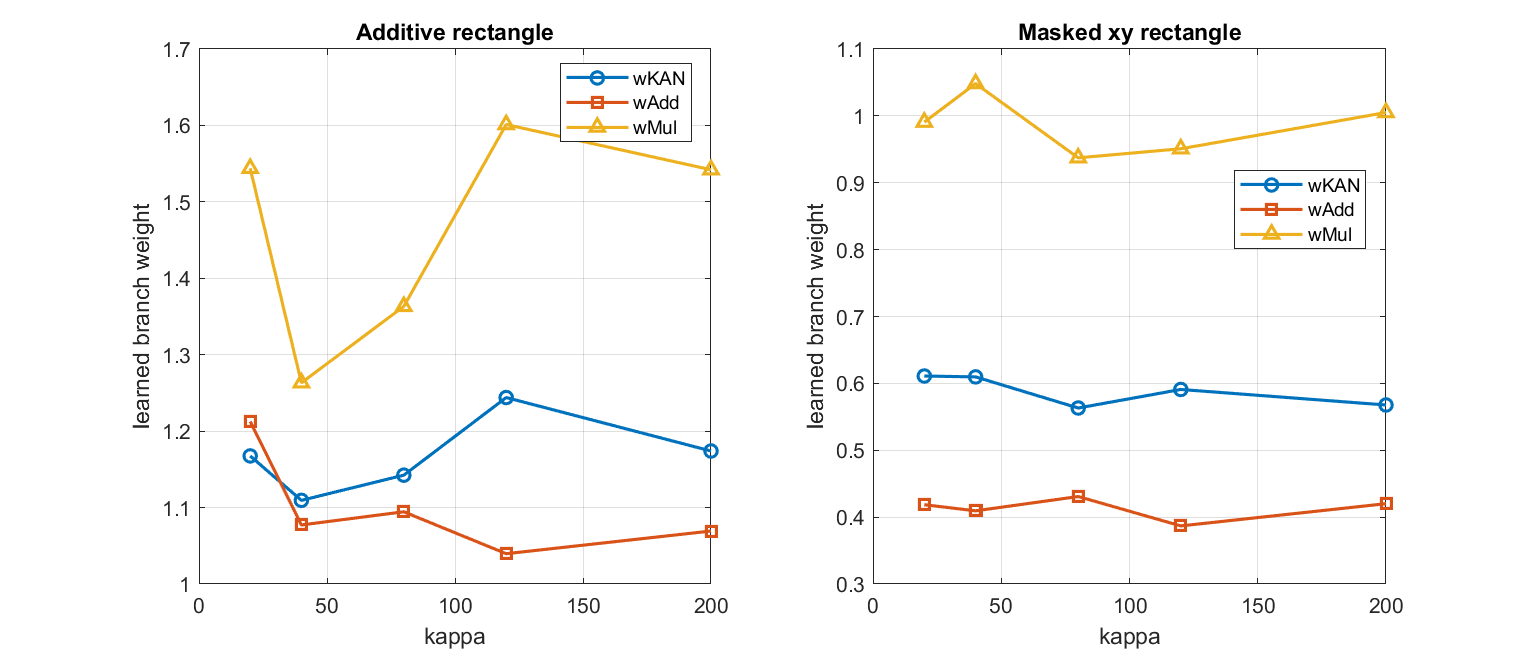}
    \caption{Learned $w_{\mathrm{KAN}}$, $w_{\mathrm{add}}$, and $w_{\mathrm{mul}}$.}
  \end{subfigure}
  \hfill
  \begin{subfigure}{1.2\linewidth}
    \centering
    \includegraphics[width=\linewidth]{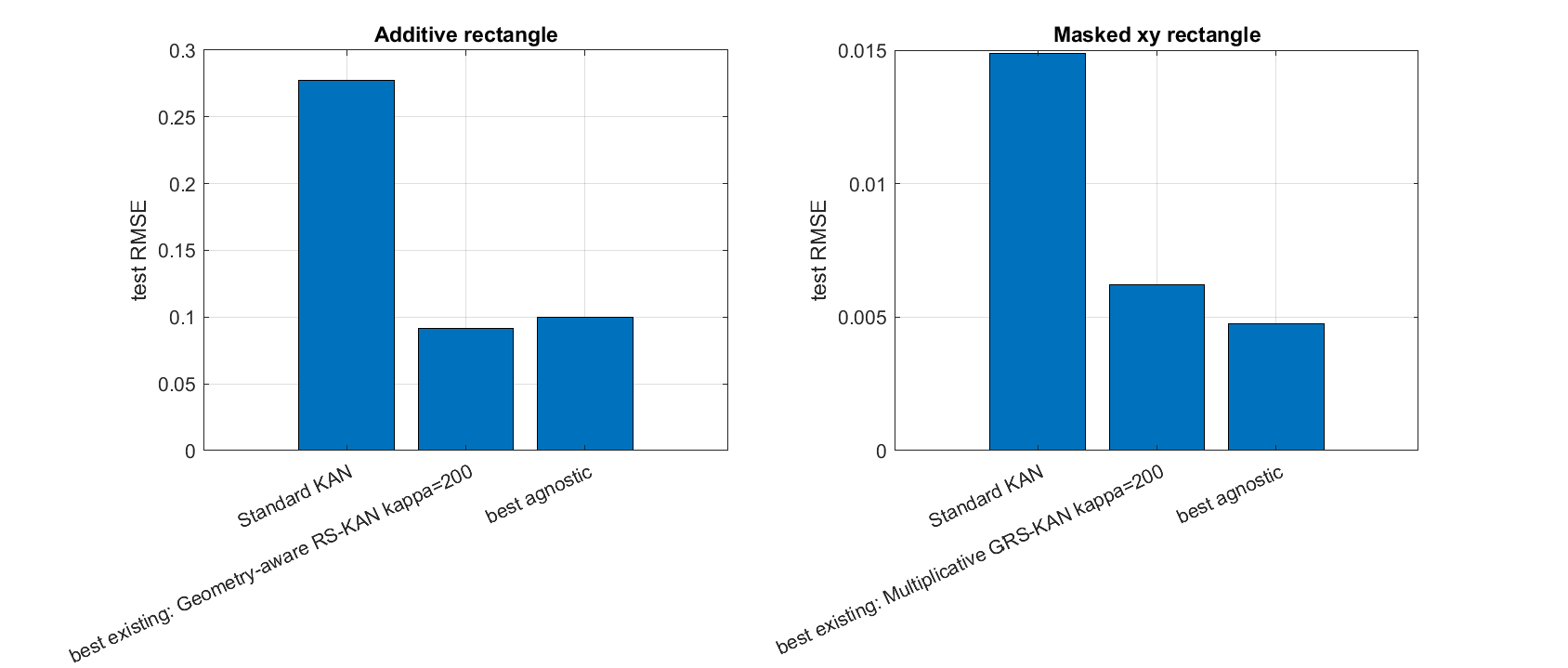}
    \caption{Best test RMSE comparison against specialized models.}
  \end{subfigure}
  \caption{Agnostic GRS-KAN branch selection on rectangular tasks.}
  \label{fig:agnostic-rect}
\end{figure}

Figure~\ref{fig:agnostic-rect} summarizes the learned branch selection. For the additive
discontinuity, all three weights remain active, so the model finds a mixed representation
rather than a clean additive-only structure. For the masked $xy$ target, $w_{\mathrm{mul}}$
dominates the geometry-aware part, matching the expected multiplicative form. The RMSE
comparison shows that the agnostic model is competitive and even improves over the
specialized multiplicative model on the masked product, although it does not beat the
specialized additive GRS-KAN on the additive discontinuity.

For the masked $xy$ task, the model learned $w_{\mathrm{mul}} \approx 1.0$, correctly identifying that the multiplicative structure is most appropriate. For the additive discontinuity, all three branches contributed, reflecting the additive jump nature.

\subsection{Sanity Check: Unconstrained Product}

As a validation, the agnostic GRS-KAN was tested on the unconstrained target $f(x,y)=xy$, which has no rectangular support or jump.

\begin{table}[H]
  \centering
  \caption{Unconstrained $xy$: KAN vs. agnostic GRS-KAN.}
  \begin{tabular}{lrrrrrr}
    \toprule
    Model & $\kappa$ & Test RMSE & Params & $w_{\mathrm{KAN}}$ & $w_{\mathrm{add}}$ & $w_{\mathrm{mul}}$ \\
    \midrule
    Full KAN & -- & 0.00720 & 255 & -- & -- & -- \\
    Pruned KAN & -- & 0.000516 & 102 & -- & -- & -- \\
    Symbolic snapped KAN & -- & $2.63\times 10^{-17}$ & 9 & -- & -- & -- \\
    Agnostic GRS-KAN & 40 & 0.000745 & 309 & 1.25 & $4.94\times 10^{-7}$ & 0.00188 \\
    \bottomrule
  \end{tabular}
  \label{tab:agnostic-xy}
\end{table}

\begin{figure}[H]
  \centering
  \begin{subfigure}{0.6\linewidth}
    \centering
    \includegraphics[width=\linewidth]{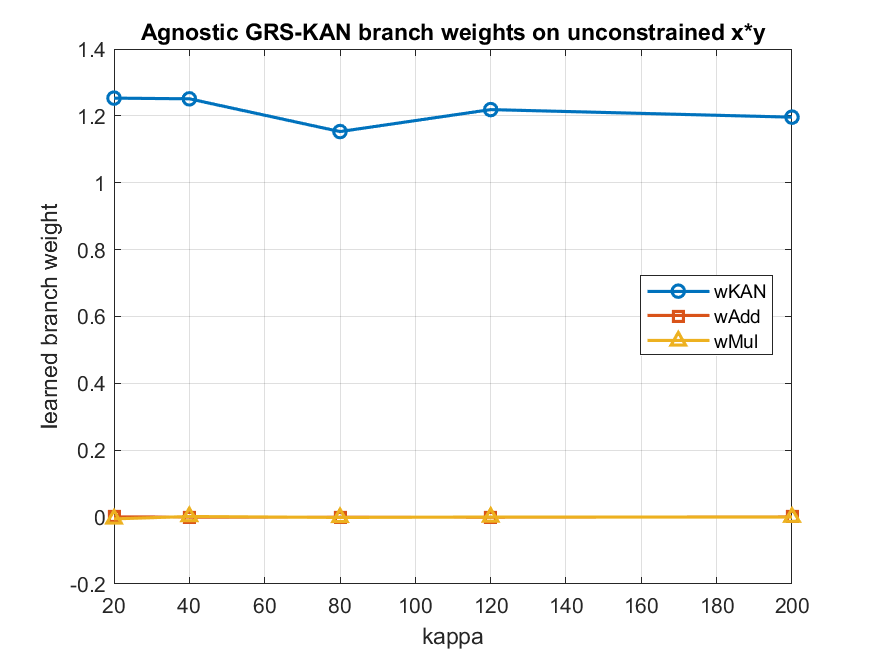}
    \caption{Learned branch weights for an unconstrained target.}
  \end{subfigure}
  \hfill
  \begin{subfigure}{0.6\linewidth}
    \centering
    \includegraphics[width=\linewidth]{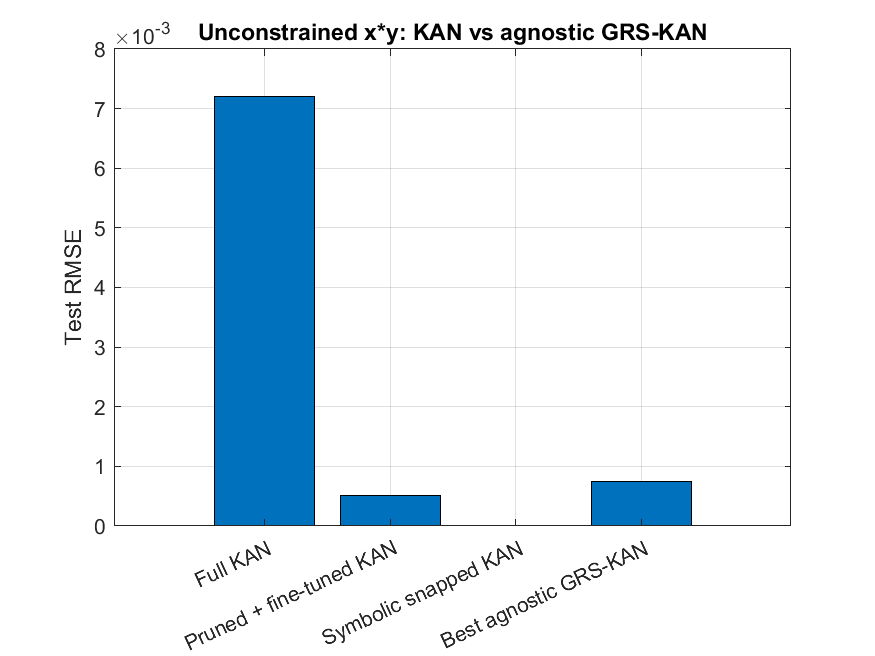}
    \caption{KAN and agnostic GRS-KAN test RMSE comparison.}
  \end{subfigure}
  \caption{Agnostic GRS-KAN on unconstrained $xy$. The geometry branches are effectively suppressed.}
  \label{fig:agnostic-xy}
\end{figure}

Figure~\ref{fig:agnostic-xy} is a negative-control experiment for the geometry-aware
architecture. Since the target has no rectangular constraint, the desired behaviour is
to suppress the geometry branches. The learned weights show exactly this:
$w_{\mathrm{add}}$ and $w_{\mathrm{mul}}$ are close to zero, while $w_{\mathrm{KAN}}$
remains large. The agnostic model therefore falls back toward a pure KAN representation,
although the pruned and symbolically snapped KANs remain more compact for this
unconstrained smooth target.

The model learned $w_{\mathrm{add}} \approx 0$ and $w_{\mathrm{mul}} \approx 0$, with $w_{\mathrm{KAN}}$ remaining large. This is the desired behavior: when no geometric prior is needed, the model falls back to a pure KAN branch and suppresses the geometry branches.

\section{Discussion}

The experiments support three main conclusions.

\paragraph{1. Standard KAN reproduces expected sparse structures.}
For the KAN paper toy example, sparsification and pruning reduced the model from $[2,5,1]$ to $[2,1,1]$. For $xy$, pruning gave $[2,2,1]$ and symbolic snapping recovered the exact product. This verifies that our MATLAB baseline correctly implements the core KAN mechanisms.

\paragraph{2. Explicit geometry improves performance on structured problems.}
For the rectangular additive discontinuity, the targeted additive GRS-KAN reduced test RMSE by $67\%$ and boundary RMSE by $61\%$ compared to standard KAN. For the masked product, the multiplicative GRS-KAN achieved $58\%$ and $50\%$ reductions respectively.

\paragraph{3. Learnable branch weights enable automatic structure discovery.}
The agnostic GRS-KAN correctly identified the multiplicative structure for the masked product ($w_{\mathrm{mul}} \approx 1.0$) and suppressed geometry branches for the unconstrained product ($w_{\mathrm{add}}, w_{\mathrm{mul}} \approx 0$). This capability is valuable when the appropriate functional form is not known a priori.

\section{Conclusions and Future Work}

We have introduced the Geometry-aware R-Structured Kolmogorov--Arnold Network (GRS-KAN), a hybrid architecture that integrates analytical implicit geometry, represented by R-functions, into Kolmogorov--Arnold Networks. The proposed framework enables:

\begin{itemize}
\setlength{\itemsep}{-2pt}
\setlength{\parskip}{0pt}
\setlength{\parsep}{0pt}
\item Analytical representation of geometric regions and Boolean constraints within differentiable neural architectures,
\item Closed-form construction of differentiable region indicators with analytical gradients,
\item Three complementary GRS-KAN architectures (targeted additive, targeted multiplicative, and agnostic) for incorporating different types of analytical geometric priors,
\item Automatic structure discovery through learnable structure-selection parameters.
\end{itemize}

Numerical experiments on rectangular discontinuity and masked product benchmarks demonstrate that incorporating analytical geometric priors significantly improves both prediction accuracy and model interpretability compared with standard KANs.

Although this paper employs simple analytical geometries such as rectangles and circles as illustrative examples, the proposed framework is applicable to arbitrary implicit regions represented by R-functions, including non-convex domains, disconnected regions, and Boolean combinations of multiple constraints.

Future work will investigate incorporating complex analytical R-function representations of engineering design spaces and feasible regions into GRS-KAN. This would enable the integration of analytical geometric priors derived from process constraints, quality specifications, and other engineering models, thereby extending the present framework from simple benchmark geometries to general implicit domains.

\bibliographystyle{plain}
\bibliography{references}

\end{document}